\documentclass[10pt]{article} 
\usepackage[preprint]{tmlr}

\usepackage{amsmath,amsfonts,bm}

\def\Eqref#1{equation~\ref{#1}}
\def\Eqref#1{Equation~\ref{#1}}

\def\1{\bm{1}}

\DeclareMathAlphabet{\mathsfit}{\encodingdefault}{\sfdefault}{m}{sl}
\SetMathAlphabet{\mathsfit}{bold}{\encodingdefault}{\sfdefault}{bx}{n}

\usepackage{hyperref}
\usepackage{url}
\usepackage{wrapfig}
\usepackage{amsmath}
\usepackage{subfig}
\usepackage{graphicx}
\usepackage{mathtools}
\usepackage{float}

\newcommand{\odesolve}{\texttt{ODESolve}}
\renewcommand{\mid}{\>|\>}
\newcommand{\map}{\text{MAP}}
\newcommand{\dd}{\mathrm{d}}

\title{Uncertainty and Structure in Neural Ordinary Differential Equations}

\author{\name Katharina Ott\\
      \addr Bosch Center for Artificial Intelligence, Renningen\\
      University of Tübingen
      \AND
      \name Michael Tiemann\\ 
      \addr Bosch Center for Artificial Intelligence, Renningen
      \AND
      \name Philipp Hennig\\
      \addr University of Tübingen \\
      MPI for Intelligent Systems Tübingen
      }

\begin{document}

\maketitle

\begin{abstract}
Neural ordinary differential equations (ODEs) are an emerging class of deep learning models for dynamical systems. They are particularly useful for learning an ODE vector field from observed trajectories (i.e., inverse problems). We here consider aspects of these models relevant for their application in science and engineering. Scientific predictions generally require structured uncertainty estimates. As a first contribution, we show that basic and lightweight Bayesian deep learning techniques like the Laplace approximation can be applied to neural ODEs to yield structured and meaningful uncertainty quantification. But, in the scientific domain, available information often goes beyond raw trajectories, and also includes mechanistic knowledge, e.g.,~in the form of conservation laws. We explore how mechanistic knowledge and uncertainty quantification interact on two recently proposed neural ODE frameworks---symplectic neural ODEs and physical models augmented with neural ODEs. In particular, uncertainty reflects the effect of mechanistic information more directly than the predictive power of the trained model could. And vice versa, structure can improve the extrapolation abilities of neural ODEs, a fact that can be best assessed in practice through uncertainty estimates. Our experimental analysis demonstrates the effectiveness of the Laplace approach on both low dimensional ODE problems and a high dimensional partial differential equation.
\end{abstract}

\section{Introduction}
Ordinary differential equations (ODEs) are a powerful tool for modelling dynamical systems.
If the dynamics of the underlying system are partially unknown and only sampled trajectories are available, modelling the vector field poses a learning problem.
One option is to parametrize the right-hand side of an ODE with a neural network, commonly known as a neural ODE \citep{chen2018neural}.
Yet, even if the exact parametric form of the underlying dynamics is unknown, we often have some structural information available.
Examples include partial knowledge of the parametric form, or knowledge of symmetries or conservation laws observed by the system.
This structural knowledge can be incorporated in the neural ODE architecture.
For example, \citet{zhong2020symplectic, zhong2020dissipative} encode Hamiltonian dynamics and dissipative Hamiltonian dynamics into the structure of the neural ODE using Hamiltonian neural networks \citep{greydanus2019hamiltonian}.
\citet{yin2021augmenting} exploit knowledge about the underlying physical model by augmenting a known parametric model with a neural ODE.
Both approaches provide a more informative prior on the network architecture giving the models superior extrapolation behavior in comparison to plain neural ODEs.
This kind of structure helps, but does not completely remove the need for training data.
When there is just not enough data available to identify the system,
meaningful predictive uncertainties are crucial.
Structured uncertainty can help quantify the benefit arising from structural prior knowledge.
Bayesian inference provides the framework to construct and quantify this uncertainty.
Generally, a fully Bayesian approach can be slow or infeasible in the context of deep learning but the Laplace approximation \citep{mackay1992practical, ritter2018a, daxberger2021laplace} enables the use of approximate Bayesian methods in deep learning.
The advantage of the Laplace approximation is that it is applied post-training, which avoids adding additional complexity to the model during training, and the model maintains the predictive performance of the \textit{maximum a posteriori} (MAP) trained model.

In this work, we apply the Laplace approximation to neural ODEs to obtain uncertainty estimates for ODE solutions and the vector field. 
Doing so is not a straightforward application of previous works on Laplace approximations, because of the nonlinear nature of the ODE solution.
We then demonstrate that the Laplace approximated neural ODEs provide meaningful, structured uncertainty, which in turn provides novel insight into the information provided by mechanistic knowledge.
Specifically, the uncertainty estimates inform us how confident we can be in the model's extrapolation.

\begin{wrapfigure}{r}{0.5\textwidth}
	\begin{center}
	\includegraphics[width=0.5\textwidth]{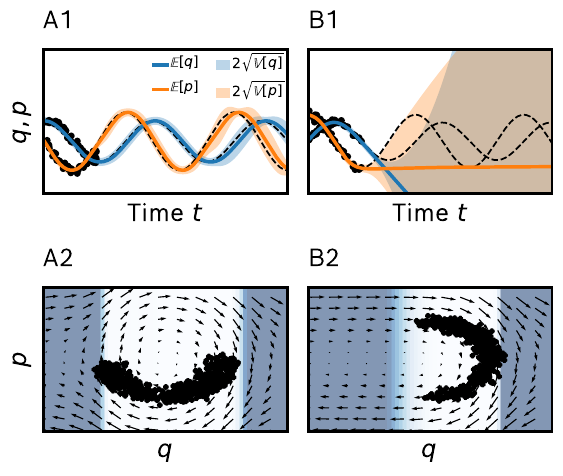}
	\caption{
		\textit{Structure of training data impacts model uncertainty.}
		Training of a Hamiltonian neural ODE on two different datasets (C--D), with Laplace-approximated uncertainty.
		$q, p$ describe position and momentum of the particle.
		(A1--B1) show trajectories for each dataset, solid lines correspond to the MAP output.
		(A2--B2) Vector field recovered by the model.
		Background color indicates the uncertainty estimates (bright means certain, dark means uncertain).
	}
	\label{fig:figure1}
	\end{center}
\end{wrapfigure}

As an example for intuition, we use a Hamiltonian neural ODE (for details see Section~\ref{sec:hamiltonian_nodes}), trained on data generated from the harmonic oscillator.
We apply the Laplace approach to find uncertainty estimates for the trained model.
The harmonic oscillator (without friction) is the textbook case of an energy-conserving system, and Hamiltonian neural ODEs capture precisely this conservation property.
We use two slightly different datasets, the only difference being they are shifted by a quarter period (corresponding to a rotation by 90 degrees in phase space see Figure~\ref{fig:figure1} (A2--B2)).
For the first dataset the solution in the extrapolation regime follows the true solution closely (see Figure~\ref{fig:figure1} (A1)).
This behavior is reflected in the low uncertainties around the solution and the large region of high confidence in the vector field (Figure~\ref{fig:figure1} (A1) and (A2)).
On the other hand, for the second dataset the extrapolation diverges quickly from the true solution, which is reflected in the high uncertainty in the extrapolation region.
The reason for this difference in model precision is that the architecture captures the dependence on $p$ explicitly, which can be exploited in one case, but not in the other. 
The same raw number of data points can thus be more or less informative, depending on where they lie in phase space.

In dynamical systems, the trajectory may leave the data domain eventually. Even if the combination of structural prior and dataset is sufficient to provide good extrapolations close to the training conditions, small changes in the initial conditions can eradicate this ability.
Without uncertainty estimates (or knowledge of the true dynamics), it is then difficult to judge the validity of the extrapolation.

\section{Technical Background}
This briefly introduces neural ODEs, and the Laplace approximation.
Section \ref{sec:uncertain_nodes} then combines these concepts to find uncertainty estimates for neural ODEs.
\subsection{Neural ODEs}
\label{sec:neural_ode}
Neural ODEs are differential equations where the right-hand side, the vector field, is parametrized by a neural network $f_{\theta}(z)$ with weights $\theta$
\begin{equation}
\frac{\mathrm{d}z}{\mathrm{d}t} = z' = f_{\theta} (z),\quad t\in[t_0, t_N],\quad z(t_0)=z_0.
\label{eq:node}
\end{equation}
In general, neural ODEs cannot be solved analytically, and a numerical scheme has to be employed, here denoted by \odesolve{} (Runge-Kutta methods \citep{hairer1993solving} are common concrete choices):
\begin{equation}
z(t_n) = \odesolve{}(f_{\theta}, z_0, [t_0, t_n]).
\end{equation}
$t_n$ denotes the time point of a specific output.

We consider regression tasks $\mathcal{D} = (z_0, t_0, \{t_n, y_n\}_{n=1}^N)$, where $y_n$ defines the outputs and $t_n$ the corresponding points in time.
This translates to an empirical risk minimization task of the form
\begin{equation}
	L_{\mathcal{D}} = \sum_{(x_n, y_n) \in \mathcal{D}} \ell(\odesolve{}(f_{\theta}, x_n), y_n),
\end{equation}
where $\ell$ is a standard loss function (i.e.,~square loss for regression tasks).
We use $x_n = \{z_0, [t_0, t_n]\}$ to denote the inputs for the \odesolve{}.

\subsection{Laplace Approximation}
\label{sec:laplace}
To equip neural ODEs with lightweight, post-training uncertainties, we focus on the Laplace approximation \citep{mackay1992practical}, a relatively old but recently resurgent method \citep{ritter2018a, kristiadi2020being,immer2021improving,daxberger2021laplace}.
We briefly review how this approach constructs an approximate posterior $p(\theta \mid \mathcal{D})$ over the weights of the net, and from there a predictive distribution $p(y\mid x, \mathcal{D})$ over its outputs.

The prior on the weights is commonly assumed to be Gaussian $p(\theta) = \mathcal{N} (\theta \mid 0, \sigma_0^2 \mathbf{I})$.
The variance of the prior $\sigma_0^2$ can be determined from weight decay, but we tune $\sigma_0^2$ by maximizing the marginal likelihood \citep{daxberger2021laplace}.
Since we want to compute the posterior over the weights, we need to find the likelihood $p(\mathcal{D} \mid \theta)$.
For mean squared error loss function the loss can be interpreted as the log likelihood $L(\theta) \propto -\log p(\mathcal{D} \mid \theta)$ ($L$ denotes the loss without a regularization term).
The posterior probability is given by $p(\theta \mid \mathcal{D}) = p(\mathcal{D} \mid \theta)p(\theta)/Z$; however we still need to calculate the normalization constant $Z$, i.e., the integral over prior and likelihood.
Since this integral for $Z$ is intractable, we approximate the posterior distribution with a Gaussian.
We start with the assumption that the final weights for a neural network trained with weight decay are a mode of the posterior, i.e., the maximum a-posteriori (MAP) estimate $\theta_{\map}$.
By Taylor-approximating the negative log posterior of the weights around the MAP estimate up to second order, we arrive at a Gaussian approximation for the weight posterior $p(\theta \mid \mathcal{D}) \approx \mathcal{N} (\theta \mid \theta_{\map}, \Sigma) \eqqcolon q(\theta).$
The variance is given by the inverse Hessian of the negative log posterior 
$\Sigma = (-\nabla^2 \log p(\theta \mid\mathcal{D} )|_{\theta_{\map}})^{-1}= (\nabla_{\theta}^2 \mathcal{L}(\theta)|_{\theta_{\map}} - \sigma_0^2 \mathbf{I})^{-1}.$
An important feature of the Laplace approximation is that it leaves the point estimate $\theta_{\map}$ untouched.
While this can be criticized from a conceptional perspective (the mean of the true posterior is not generally identical with its mode), it is convenient from a practical standpoint, since $\theta_{\map}$ is precisely the estimate practitioners spend much time getting right with elaborate training procedures.
Calculating the full Hessian is costly, and the Laplace approximation expects the Hessian to be positive semi-definite. A common way to address both issues is to approximate the Hessian with the generalized Gauss-Newton matrix (GGN) which only involves the evaluation of the Jacobian with respect to the weights \citep{daxberger2021laplace}.

\section{Laplace Approximation for Neural ODEs}
\label{sec:uncertain_nodes}

Where neural ODEs are used in the scientific domain, the model output should include quantified uncertainties, assessing the reliability of predictions.
In this section we extend the Laplace approach to neural ODEs and introduce how to compute uncertainty estimates for neural ODEs.
For our implementation we extend \texttt{laplace-torch} \citep{daxberger2021laplace} to neural ODEs.

Given the approximate Hessian, we can calculate the predictive distribution for new data via
\begin{equation}
	\label{eq:1}
	p(y\mid x, \mathcal{D}) = \int p(y\mid\odesolve{}(f_{\theta}, x)) q(\theta) \dd \theta.
\end{equation}
Since this integral is analytically intractable, some form of approximation has to be applied.
Thus, applying the Laplace approximation to the neural ODE architecture poses a few technical challenges. 
In particular, the predictive distribution of \odesolve{} can be computed by sampling and linearization.
Below, we discuss both options, and argue that linearization is favorable.
Finally, we show how to find the predictive distribution of the vector field.

\paragraph{Sampling the Network Weights}
A first way to approximate \Eqref{eq:1} is to sample the weights of the neural net from the posterior distribution
\[p(y\mid x, \mathcal{D}) = \frac{1}{N} \sum_{i= 1}^{N}p(y\mid\odesolve{}(f_{\theta_i}, x)),\]
for $\theta_i \sim q(\theta)$.
The neural ODE is then solved for each of these weight configurations.
This requires solving a neural ODE repeatedly, for each perturbed vector field.

\paragraph{Linearizing the ODESolve}
\citet{farquar2020liberty, khan2019approximate} suggest linearizing the neural network with respect to the weights.
In this case sampling is no longer necessary since the predictive distribution can be obtained in closed form.
For neural ODEs, this corresponds to linearizing the entire \odesolve{} around the MAP with respect to the parameters
\begin{equation}
  \odesolve{}(f_{\theta}, x) \approx \odesolve{}(f_{\theta_{\map}}, x) + J_{\theta_{\map}}(x) (\theta - \theta_{\map}),
\end{equation}
where $J_{\theta_{\map}}$ is the Jacobian of $\odesolve{}$ with respect to the parameters $[J_{\theta_{\map}}]_{i,j} = \frac{\partial \odesolve{}_i}{\partial \theta_j} (\theta_{\map}, x)$.
The Jacobian of \odesolve{} is computed using automatic differentiation functionalities.
The predictive distribution can now be obtained in closed form
\begin{equation}
p(y \mid x, \mathcal{D}) \approx \mathcal{N}(y\mid \odesolve{}(f_{\theta_{\map}}, x), J_{\theta_{\map}}(x)^T \Sigma J_{\theta_{\map}}(x) + \sigma^2 I),
\end{equation}
where $\sigma^2$ is the variance of the observation noise (for more details we refer to \cite{daxberger2021laplace}).

Given the two approaches to find the predictive distribution of the \odesolve{}, which one is preferable?
Sampling, in combination with GGN approximation of the Hessian, does not provide useful uncertainties, possibly due to the mismatch between the approximation and true Hessian.
Additionally, \citet{immer2021improving} show that the GGN implicitly assumes linearization, so sampling may not provide additional benefits.
For a comparison of the sampling and the linearization approach we refer to Figure \ref{sfig:sample} in the Supplementary Material.
Thus, we use a linearization of the \odesolve{} to approximate the predictive distribution.

\paragraph{Linearizing the Vector Field}
The key to understanding the interplay between data, model structure and extrapolation quality and uncertainty lies in understanding which parts of the vectorfield have been identified through the available data and the model structure.
Although linearization provides closed-form predictive distributions for the outputs of the \odesolve{}, it does not provide uncertainties for the vector field $f_{\theta}$.
Instead of linearizing the \odesolve{}, another option is to just linearize the vector field with respect to the parameters.
However, in this case, the GGN approximation is no longer equal to the Hessian of the linearized model.
Linearizing the vector field allows to calculate the predictive distribution for the vector field in closed form
\[p(z' \mid z, \mathcal{D}) \approx \mathcal{N}\left(z'\mid f_{\theta_{\map}}(z), J(z)^T \Sigma J(z)\right),\]
where $J$ is in this case the Jacobian  of the vector field $f$ with respect to the parameters.
Here $y$ corresponds to the output of the vector field.

\subsection{Comparison to HMC}
\label{ssec:comparison}
\begin{figure*}[t!]
	\includegraphics[width=\textwidth]{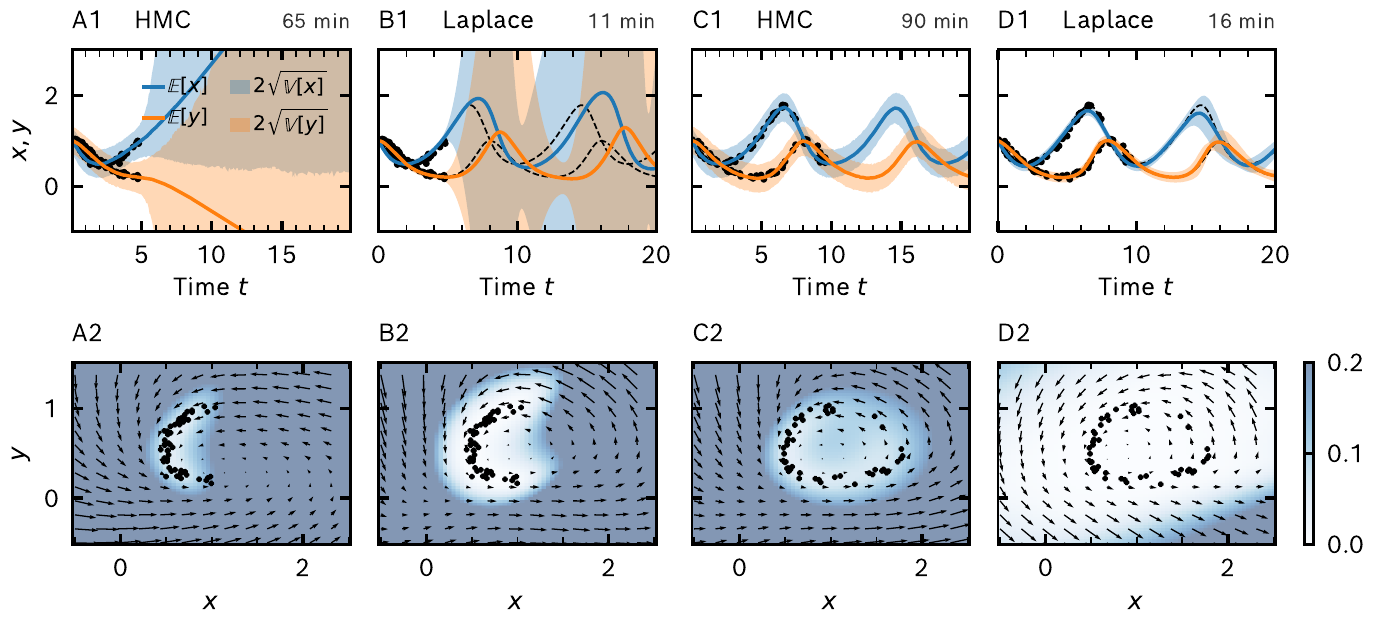}
	\caption{
		\textit{Comparison of HMC and Laplace approximation on the Lotka Volterra dataset.}
		(A1-D1) Trajectories (lines) with uncertainty estimates.
		(A2-D2) Vector field with uncertainty estimates.
		The background color indicates the norm of the uncertainty estimates.}
	\label{fig:lv}
\end{figure*}
For an empirical evaluation, we here compare the quality of uncertainty estimates provided by the Laplace approximation to Monte Carlo sampling.
Specifically, we use Hamiltonian Monte Carlo (HMC) with no U-turn sampling (NUTS) (Section~\ref{ssec:comparison_hmc} in the Supplements).
To compare HMC and the Laplace approach, we introduce two datasets based the Lotka Volterra equations (see \Eqref{eq:lv} in the Appendix, for additional results on the pendulum dataset we refer to Section \ref{sec:sup_exp_hmc} in the Supplementary Material).
\emph{dataset-half-cycle} only provides data for half a cycle whereas \emph{dataset-full-cycle} provides data about the entire cycle.
For \emph{dataset-half-cycle}, both HMC and Laplace are uncertain about the extrapolation outside data domain (see Figure~\ref{fig:lv} (A1) and (B1)), in contrast to \emph{dataset-full-cycle} where both models extrapolate accurately with high confidence.
This is also reflected in the uncertainty estimates for the vector field (see Figure \ref{fig:lv} (E--F)).
Notably, the mean extrapolation for HMC and Laplace behave differently as the Laplace extrapolation corresponds to the MAP whereas the HMC extrapolation corresponds to the mean of the posterior.
Furthermore, the uncertainty estimates of the Laplace approximation decrease whenever the MAP solution returns to data domain in phase space.
This is likely a shortcoming of the linearization approach, which neglects that the uncertainties should add up over the course of the \odesolve{}.
As indicated by high uncertainty estimates, the similarity between the MAP extrapolation and the true solution is just coincidence as there are other weight configurations which do not extrapolate well.

One of the biggest advantages of the Laplace approximation is that is computationally much cheaper than HMC (see Figure \ref{fig:lv} for the runtime of each experiment).
In our experiment we choose a sufficiently small number of weights to enable HMC inference, however already doubling the network weights proved to be infeasibly slow for HMC (surpassing the allocated runtime of 24 h) whereas training and inference time for the Laplace approximation barely increased.
We find that the Laplace approximation is slightly inaccurate in its uncertainty estimates due to the linearization approach, but the superior runtime performance justifies the output quality.

\section{Structure Interacts with Uncertainty}

Recent approaches \citep{yin2021augmenting,zhong2020symplectic} in using neural ODEs for dynamics model learning aim to improve the modelling capabilities of neural ODEs by including structural knowledge about the dynamics.
We will see that, for such models, even small shifts in the dataset can disproportionally change the prediction of the neural ODE (see Figure \ref{fig:figure1}), because the structural information causes complicated interactions with the identifiability, and hence, uncertainty of the phase space.
Experiments below show that a small change in dataset and structural information can lead to a wide range of results, but that the Laplace approximation serves as a reliable tool to characterize and visualize this issue through the notion of uncertainty.

\subsection{Hamiltonian Neural ODEs}
\label{sec:hamiltonian_nodes}
Dynamical systems of practical concern often come with information about the underlying physics.
One option to introduce knowledge about the underlying dynamics of a system is via conservation laws, e.g., the Hamiltonian equation of motion. For energy-conserving systems, the dynamics obey the relation
\begin{equation}\label{eq:hamiltonian}
	\dot{q} = \frac{\partial H(p,q)}{\partial p}, \qquad \dot{p} = -\frac{\partial H(p,q)}{\partial q},
\end{equation}
with a function $H(p,q)$ called the Hamiltonian, which, by the above equation, is conserved over time ($\dd H/\dd t=\nolinebreak0$).
We give a short introduction of how to construct a Hamiltonian neural ODE similar to SymODE introduced in \citet{zhong2020symplectic, zhong2020dissipative}.
For a lot of real systems it is sufficient to consider a separable Hamiltonian of the form $H(q, p) = T(p) + V(q)$.
To learn such a separable Hamiltonian from data, we use neural networks to represent $V$ and $T$ (we term this model \textbf{separable}).
To further refine the prior knowledge in the architecture, we can use that the kinetic energy $T$ is given by $T(p) = p^T M^{-1} p/2$ where $M$ is a positive definite mass matrix, a \textbf{constrained} model.
\paragraph{Uncertainty in Hamiltonian Neural ODEs}
\begin{wrapfigure}{r}{2.7in}
	\begin{center}
	\includegraphics{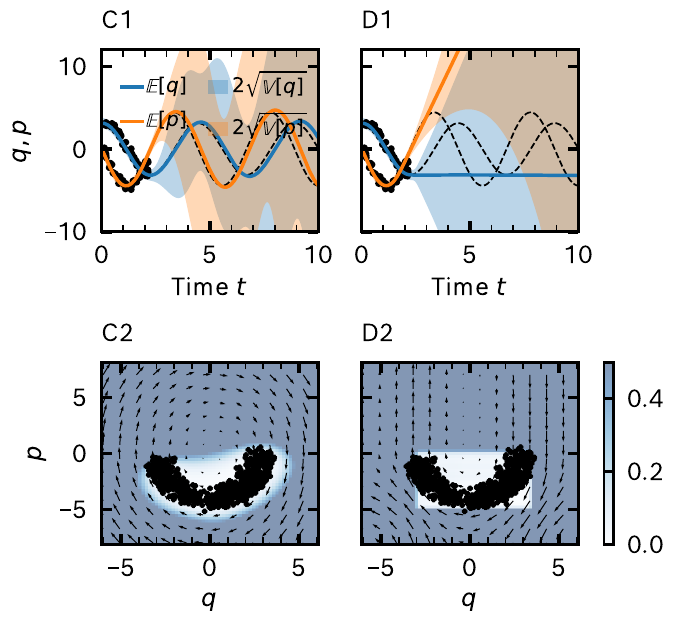}
	\caption{
			\textit{Structural information in neural ODEs affects both point and uncertainty estimates.}
			Three different architectures were used for training: naive (C1) and (C2), separable (D1) and (D2).
			(C1--D1) Solutions to the initial value problem with uncertainty estimates provided by the Laplace approximation.
			(C2--D2) Vector field of the neural ODE model.
			}
	\label{fig:uncertain_nodes}
	\end{center}
\end{wrapfigure}
We add the parameters of the matrix $M$ to the trainable parameters of our model.
In summary, there are three neural ODE models we consider in this work:
The \textbf{naive} approach of \Eqref{eq:node}, where the right-hand side, $f$, is represented by a neural network, a \textbf{separable} model and a \textbf{constrained} model.

For the following experiments, we train neural ODEs on different datasets generated from the harmonic oscillator (for experimental details and the network architecture see Supplementary Material Section~\ref{ssec:sup_hamiltonian_nodes}).
The model is only provided with data in the lower plane (i.e., \emph{dataset-lower-half}), corresponding to half a period of the particle's motion (see for example Figure~\ref{fig:uncertain_nodes}, (C1) and (C2)).
To compute uncertainty estimates for the trajectories and the vector field $f_{\theta}$ of the neural ODE we use the Laplace approximation with the linearization schemes introduced in Section \ref{sec:uncertain_nodes}.
In the region where data is available the naive model is able to fit the data.
The extrapolation for the trajectories and the vector field is close to the true dynamics, but the uncertainty estimates indicate that there is not enough information available to guarantee correct extrapolation outside the data domain (cf.~Section \ref{ssec:comparison_hmc}).

How does additional structure (i.e., a separable and a constrained neural ODE) change the extrapolation behavior on \emph{dataset-lower-half}?
Similar to the naive model, the separable model becomes uncertain outside the data domain (see Figure~\ref{fig:uncertain_nodes} (B1) and (B2)).
Although the approximation power of this model should be identical to the naive structure, and hence, an identical extrapolation quality should be achievable, there is a qualitative difference in the extrapolation of the solution and the vector field
(we attribute this to the different structure in the architecture see Appendix Section \ref{ssec:sup_training_hnods}).
The distinctive rectangular shape in the uncertainty estimates of the vector field reflects the separable structure of this model.

\begin{figure*}
	\includegraphics{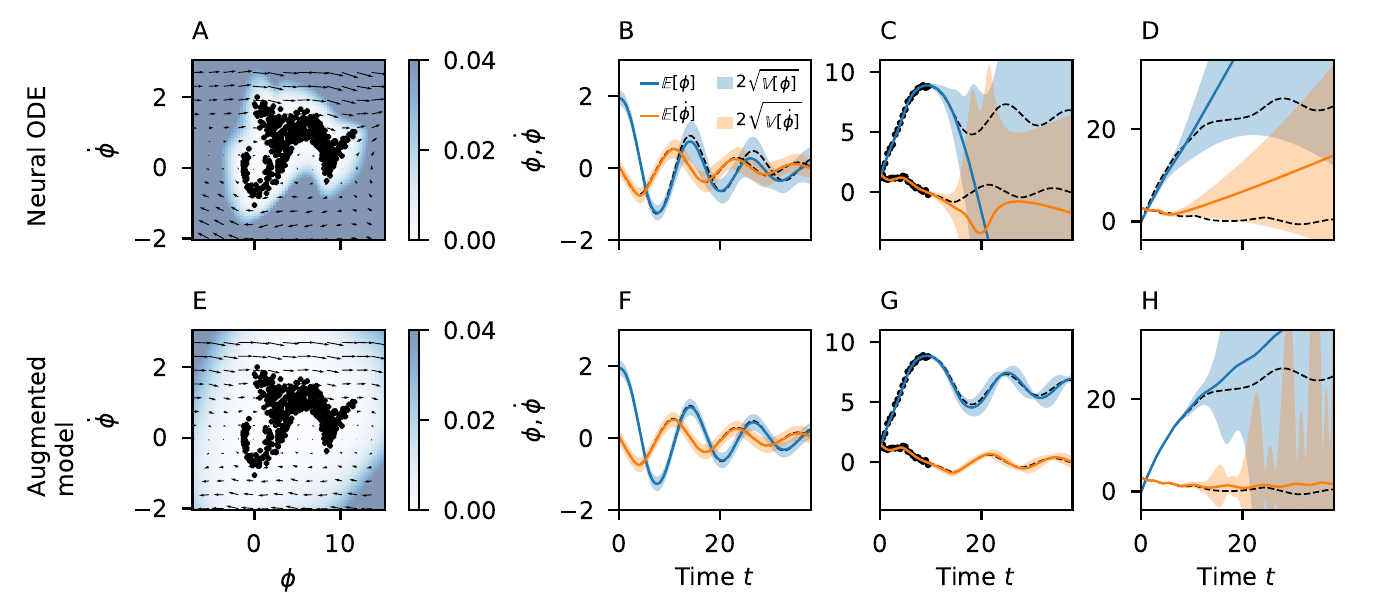}
	\caption{
		\textit{Uncertainty estimates for augmented parametric models.}
		Neural ODE and augmented parametric model trained on the damped pendulum dataset.
		(A) and (E) vector field with uncertainty estimates.
		(B -- D) and (F -- H) trajectories for different initial conditions with uncertainty estimates.
	}
	\label{fig:pendulum}
\end{figure*}

On the other hand, if we train the constrained model on \emph{dataset-lower-half}, the uncertainty estimates of the vector field exhibit a vertical band of low uncertainty around the data (Figure~\ref{fig:uncertain_nodes} (G)).
This shape is directly linked to the architecture of the model: The only trainable parameters (apart from $M$ which is a scalar), are the parameters of the potential $V_{\theta}(q)$ where learning $V_{\theta}$ depends only on the availability of data for $q$.
So wherever data identifies $V_{\theta}$ the value of the Hamiltonian $H$ is known since we use the concrete parametric form of $T(p)$.
If there is data available for some $(q,p)$ the vector field is determined in the region $(q, \mathbb{R})$, leading to uncertainty bands in the structure of the vector field.
Therefore, the constrained model is able to extrapolate with relatively low uncertainty, since the extrapolation remains within the domain where training data was available for $q$.

Conversely, if we rotate the dataset by 90 degrees (\emph{dataset-left-half}), the domain where data is available for $q$ changes significantly.
This time, however, sufficient data for $q$ is not available and therefore, the solution shows high uncertainty in the extrapolation region (Figure~\ref{fig:uncertain_nodes} (D)).
For results of the other models on \emph{dataset-left-half} see Supplementary Material Section \ref{sec:sup_exp_hamiltonian}.

These results highlight the intricate effect of mechanistic knowledge on extrapolation, and its reflection in model uncertainty:
While a particular dataset might not be sufficient for the naive model to extrapolate accurately, mechanistic knowledge in the form of conservation laws might change this.
However, seemingly benign changes in the dataset (here: a shift in phase) can substantially affect the quality of the predictions.
These effects are not always intuitive and hard to see in point predictions, but they become immediately evident when looking at the uncertainty estimates of the vector field and the solution provided by the Laplace approximation.
Given the distinct algebraic structure of the model considered in this section and the low dimensionality of the problem we can assess which data fully identifies the vector field to allow for extrapolation.
But also in more complex, high dimensional problems, where predicting the impact of the encoded structural knowledge becomes increasingly hard, the temporal evolution of the Laplace uncertainty estimates reveals the extrapolation capabilities of a model consistently. The probabilistic notion thus recommends itself when designing structural priors.

\subsection{Augmenting Parametric Models with Neural ODEs}
Instead of including knowledge of conserved quantities in the architecture of the neural ODE, another option is to consider knowledge of the parametric form of the underlying physics.
But in many cases we do not have knowledge of the full dynamics, hence \citet{yin2021augmenting} suggest augmenting a known parametric model $f_p$ with a neural network $f_{\theta}$.
The resulting dynamics of the model are given by
\begin{equation}
	\frac{\mathrm{d}z}{\mathrm{d}t} = f_{\theta} (z) + f_p (z),\quad t\in[t_0, t_N],\quad z(t_0)=z_0.
\end{equation}
To ensure that the dynamics are not dominated by the neural ODE, \citet{yin2021augmenting} suggest regularizing $f_\theta$ (for more details we refer to their paper).
We apply the Laplace approach to models trained on two datasets proposed by \citet{yin2021augmenting} -- a damped pendulum and damped wave equations.

\paragraph{Pendulum Dataset}
To show the effectiveness of the Laplace approach, we train an augmented parametric model and a plain neural ODE on a dataset describing a damped pendulum $\dd^2 \phi/\dd t^2 + \omega^2 \sin \phi + \alpha \dd \phi/\dd t = 0$, where $\phi$ is the angle, $\omega$ the frequency and $\alpha$ a damping coefficient.
For the parametric model we use frictionless pendulum dynamics where we add the frequency $\omega$ as a trainable parameter to our model.

The difference in the architecture between the two models is already evident in the uncertainty estimates of the vector field (shown in Figure~\ref{fig:pendulum} (A) and (E)).
For the augmented model, the region where the model has low uncertainty is larger than for the plain neural ODE model.
However, far away from the data both models become uncertain.
To highlight the differences between the two models we evaluate them with different initial conditions.
On the first set of initial conditions (Figure~\ref{fig:pendulum} (B) and (F)) both models are able to extrapolate accurately, which is captured by the low uncertainty estimates.
For the initial conditions in Figure~\ref{fig:pendulum} (C) and (G), both models are able to fit the data, yet the plain neural ODE model is unable to extrapolate, reflected by the large uncertainty estimates in the extrapolation regime.
There also exist initial conditions for which both models fail to extrapolate, but our experiments reveal that the Laplace approximation is able to detect these regions (see Figure~\ref{fig:pendulum} (D) and (H)).
Running HMC inference on a reduced version of this task, we observe the same behavior.
For the additional HMC results we refer to the Supplementary Material Section \ref{sec:sup_exp_hmc}. 

Given the complicated structure of the dataset and model architecture, it is \emph{a priori} unclear for which initial conditions the model extrapolates well.
Hence, it is absolutely crucial to use uncertainty estimates to assess the models' outputs.

\paragraph{High Dimensional Wave Dataset}
\begin{wrapfigure}{r}{0.6\textwidth}
	\begin{center}
	\includegraphics{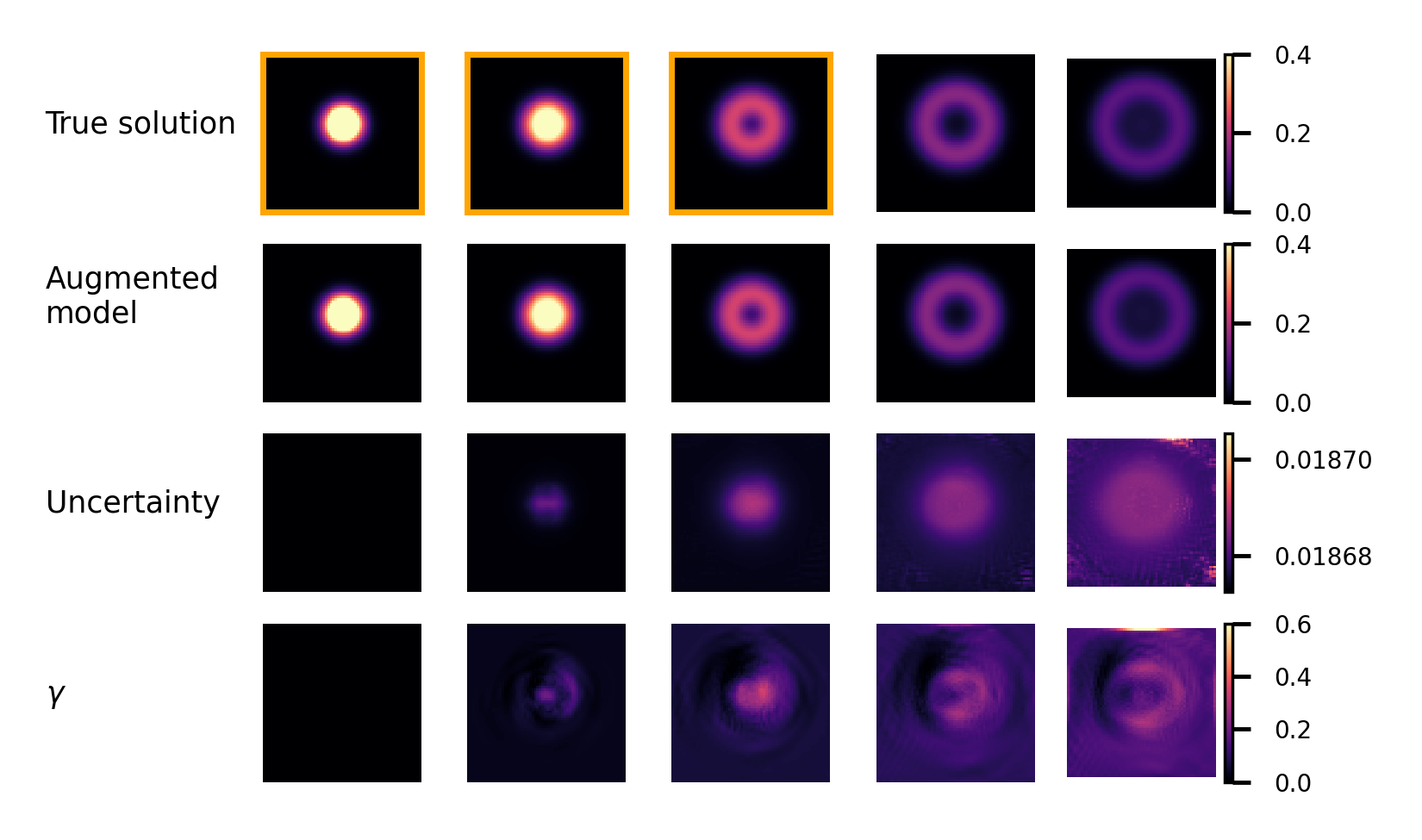}
	\caption{
		\textit{Laplace approximation applied to wave PDE.}
		Training an augmented parametric model on the damped wave equation dataset (Figure shows result for $u$ --- for the time derivative and additional results we refer to Supplementary Material \ref{sec:sup_wave_exp}).
		First three images (marked by orange frame) are part of the training data.
	}
	\label{fig:pde}
	\end{center}
\end{wrapfigure}
We show that the Laplace approximation is also applicable to high dimensional data.
In this case we use the damped wave equation as a dataset, where the wave is described by a scalar function $u$ and the wave equation is given by $\partial^2u / \partial t ^2 - c^2 \Delta u + k \partial u / \partial t = 0$.
$k$ is a damping coefficient.
The dataset consists of a 64x64 spatial discretization for $u$ and $d u/d t$ over multiple time points.
Our model consists of a parametric model for the wave equation without the damping term augmented with a neural ODE model.

The uncertainty estimates reproduce the overall structure of the wave expansion (see Section \ref{sec:sup_wave_exp} in the Appendix for different initial conditions).
Specifically, in the extrapolation regime behind the wave front the uncertainty estimates increase (see Figure \ref{fig:pde}).
To check if the uncertainty estimates are well calibrated, we compute the ratio of error and standard deviation, where the error is given by the difference between the true solution and the model's output.
We denote this ratio by $\gamma$.
The uncertainty estimates are well calibrated if $\gamma$ is close to one.
Overall, we find that the uncertainty estimates provided by the Laplace approximation are underconfident ($\gamma < 1$).
The imperfections in the uncertainty estimates are compensated by the fact the Laplace approach facilitates the computation of uncertainty estimates on such a high dimensional dataset in reasonable time---other approaches like HMC would be computationally infeasible.

\subsection{Discussion And Outlook}
\label{sec:discussion}

While our experiments suggest that the Laplace approximation produces high quality uncertainty quantification for a variety of tasks, and for various quantities, there are some numerical and computational issues to carefully consider, which we briefly discuss here.

The computationally most expensive part of the Laplace approximation is the calculation of the GGN, and especially its inverse.
But once calculated and stored, it does not have to be reevaluated for future predictions.
Since neural ODEs commonly use a relatively small network size, compared to other deep learning applications, storing the Hessian need not be an issue.
If it is, there are a few options available, like diagonalizing, only using the last layer or only considering a subnetwork to reduce the memory cost \citep{daxberger2021laplace}.
How well these methods apply to neural ODEs is left as future work.
Another issue we face is that the GGN sometimes loses its positive-semi-definiteness,
 due to numerical issues (i.e., large variance in the eigenvalues).
This can be alleviated by adding a small constant to the diagonal elements of the Hessian.
We also found that this effect is enhanced by the structure of Hamiltonian neural networks (possibly due to the derivative structure of the activation functions).
For inference, taking the Jacobian over the whole trajectory can be costly (especially for large datasets like the wave dataset).
However, in practice we are often only interested in the final output and dense sampling is not necessary.

\section{Related Work}
Neural ODEs \citep{chen2018neural} have been applied to a wide range of problems such as image classification \citep{chen2018neural, zhang2019anodev2, choromanski2020ode}, normalizing flows \citep{chen2018neural, grathwohl2018scalable}, learning dynamical systems via residual neural networks \citep{brouwer2019gru, kidger2020neural} or variational autoencoders \citep{chen2018neural, rubanove2019latent, yildiz2019ode}.

\textbf{Neural ODEs With Structure} \citet{greydanus2019hamiltonian} introduce the idea of adding a Hamiltonian structure to a neural network.
\citet{zhong2020symplectic} extend this idea to neural ODEs and \citet{zhong2020dissipative} add a term to Hamiltonian neural ODEs to model dissipative systems.
\citet{yin2021augmenting} propose to augment parametric models with neural ODEs by regularizing the neural ODE term.

\textbf{Neural ODEs With Uncertainty}
To model the latent space of a variational autoencoder \citet{yildiz2019ode} use a Bayesian neural network to describe the vector field.
Similarly, \citet{dandekar2020bayesian} train a neural ODE with a Bayesian neural network as the vector field on regression and classification tasks using Monte Carlo sampling to do inference.
\citet{yang2021bpinn} apply HMC and variational inference to physics-informed neural networks.
\citet{norcliffe2021neural} propose to use neural processes to equip neural ODEs with uncertainty estimates.
Relative to these works, ours is the first to construct and assess uncertainty quantification for neural ODEs with structured architectural priors.

Stochastic differential equations (SDEs) can be used to model the stochasticity of real-world processes.
This approach has been transferred to neural ODEs for example for training a recurrent neural network \citep{brouwer2019gru} or to do variational inference using a latent stochastic differential equation \citep{li2020scalable}.
To improve uncertainty quantification in image classification, \citet{kong2020sde} propose to use a neural SDEs and \citet{anumasa2021improving} combine a GP with a neural ODE.

\textbf{GPs for Modelling ODE Dynamics} Another approach to free-form dynamics modelling are Gaussian processes (GPs) \citep{heinonen2018learning, hedge2021bayesian}.
\citet{hedge2021bayesian} learn a posterior distribution over the vector field of the ODE.
\citet{ensinger2021symplectic} encode a Hamiltonian structure in the GP and use symplectic ODE solvers to train the model.
\citet{wang2020physics} augment incomplete physical knowledge with a GP.
\citet{ridderbusch2021learning} propose to use GPs to learn a vector field from data by using prior structural knowledge.

\section{Conclusion}

Uncertainty is always relevant in machine learning, but particularly so for the highly structured and often unintuitive prediction of dynamical systems with neural ODEs. At its most basic, uncertainty estimates tell us if the model has seen enough data to learn the dynamics.
But the position in state space, and number, of data points required for this to happen depend intricately on additional mechanistic knowledge potentially encoded in the model.
Small changes in the dataset can have a fatal effect on the extraploratory abilities of the neural ODE model. These aspects can be hard or impossible to spot from point predictions alone, yet may become obvious when uncertainty estimates are available.

To make neural ODEs a useful tool for scientific or engineering applications, it is thus crucial to make uncertainty estimates available at train- and test-time.
The experiments presented in this work suggest that Laplace approximations, with the necessary technical adaptations for this model class, can provide such uncertainties for neural ODEs at simultaneously high fidelity and at low cost.
Moreover, the uncertainty estimates are able to reflect key structural effects of mechanistic knowledge, and they thus help make neural ODEs more reliable as a tool for scientific inference.

\bibliography{main}
\bibliographystyle{tmlr}

\appendix

\section{Datasets}
\subsection{Lotka Volterra}
As one of the standard datasets to discuss ODEs, we use the Lotka Volterra system to compare the Laplace approximation to other models.
The Lotka Volterra system models the interaction between a species of predator $y$ and prey $x$.
The system is described by the following differential equations
\begin{align}
	\frac{\dd x}{\dd t} = \alpha x - \beta xy,
	\frac{\dd y}{\dd t} = \delta x y - \gamma y,  
	\label{eq:lv}
\end{align}
where $\alpha, \beta, \gamma, \delta$ are positive real parameters describing the interactions between predators and prey.
For our experiment we chose $\alpha = 2/3, \beta= 4/3, \delta=1, \gamma=1$. We add zero mean Gaussian noise with variance $\sigma^2 = 0.03$.

For \emph{dataset-half-cycle} training data is generated on the interval $t \in [0, 5]$.
For \emph{dataset-full-cycle} training data is generated on the interval $t \in [0, 10]$.
For both datasets we set $x_0 = (1, 1)$.

\subsection{Harmonic Oscillator}
The  equations of motion for a harmonic oscillator, for a particle with position $q$ and momentum $p$, are given by
\begin{equation}
\dot{q} := \frac{\dd q}{\dd t} = \frac{p}{m},\quad\dot{p} := \frac{\dd p}{\dd t} = -k x.
\label{eq:harmonic_oscillator}
\end{equation}
$k$ is the spring constant of the system, $m$ is the mass of the particle.
For the experiments we use three different datasets generated from the harmonic oscillator. 
All datasets consist of 16 trajectories with different initial conditions and each trajectory consists of 50 samples.
We add zero mean Gaussian noise with variance $\sigma^2 = 0.3$ to each dataset.
\begin{figure}[b]
	\centering
	\subfloat[]{
		\includegraphics[width=0.3\textwidth]
		{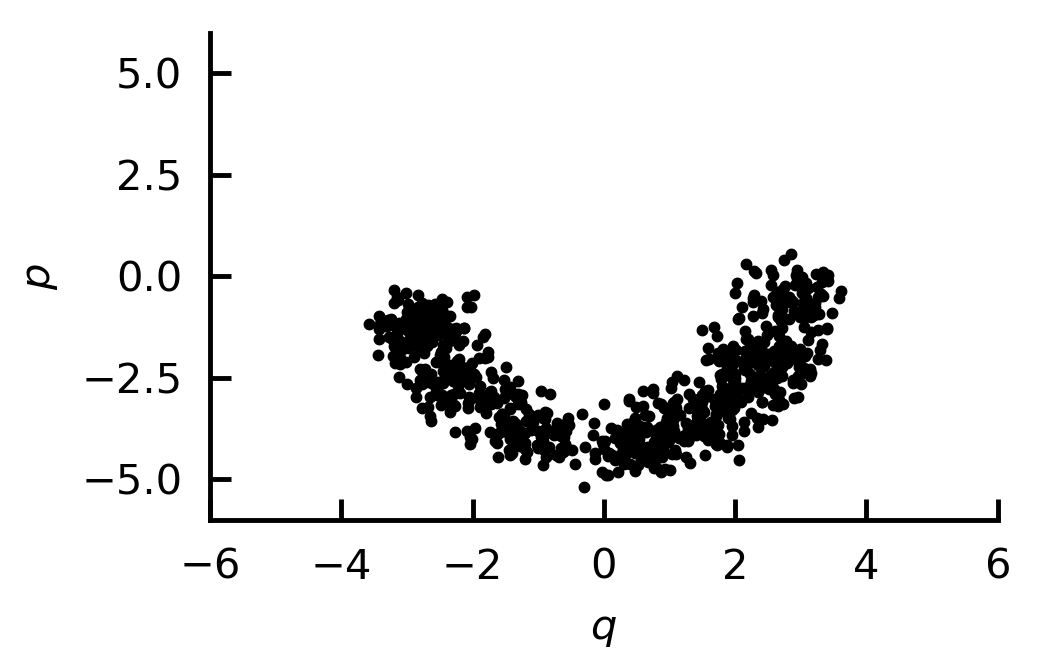}
	}
	\subfloat[]{
		\includegraphics[width=0.3\textwidth]
		{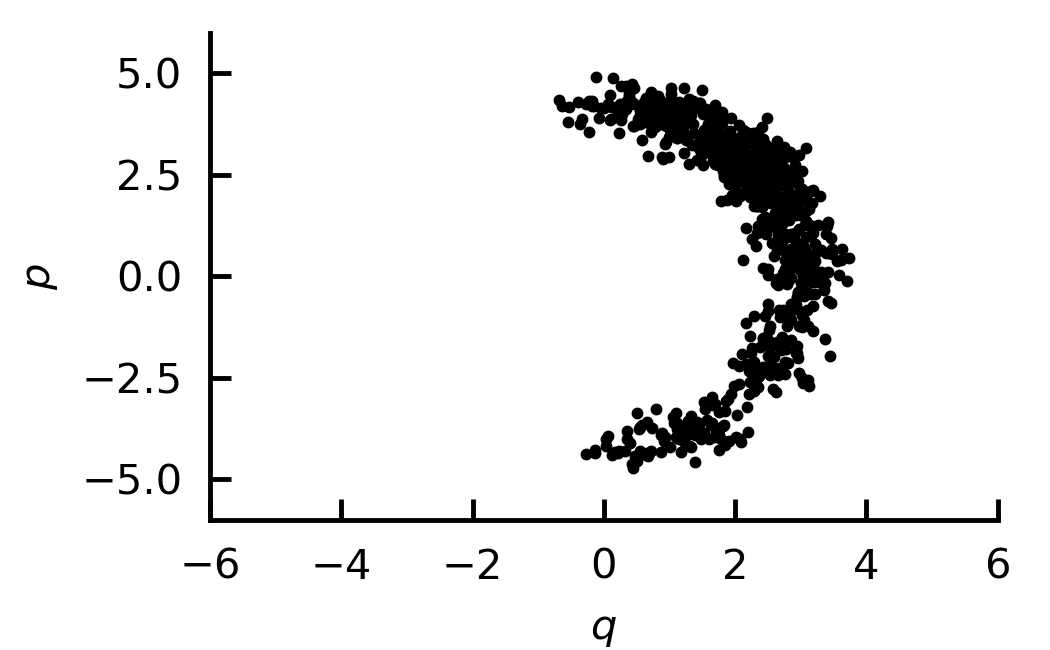}
	}	
	\subfloat[]{
		\includegraphics[width=0.3\textwidth]
		{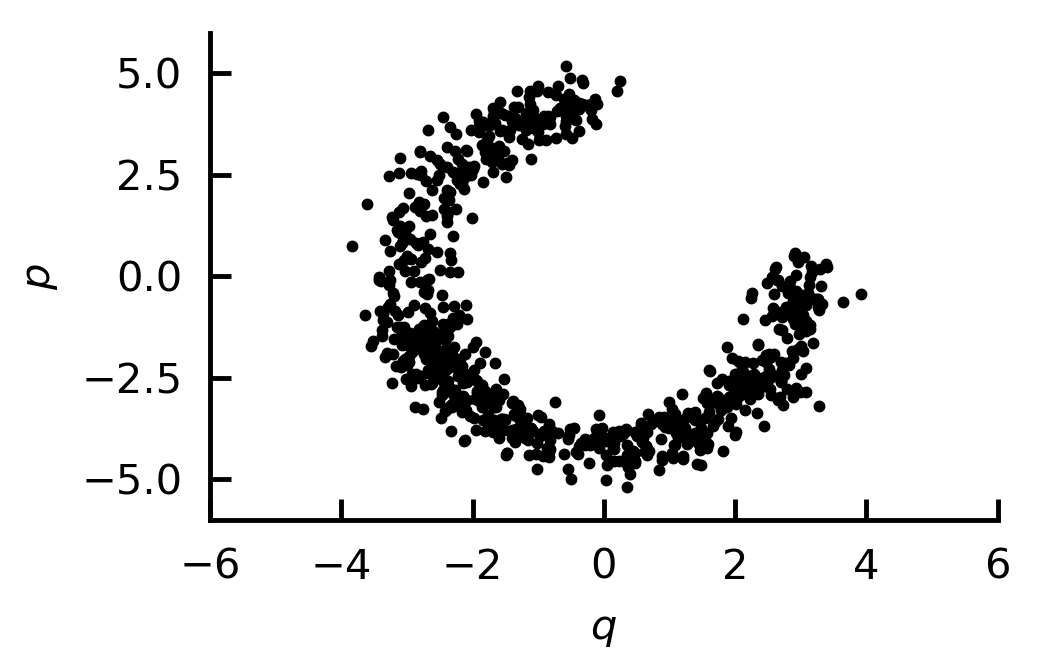}
	}
	
	\caption{Datasets used for the experiments. (a) shows \emph{dataset-lower-half}, (b) \emph{dataset-left-half}, (c) \emph{dataset-three-quarter}.}
	\label{fig:datasets}
\end{figure}

\paragraph{Dataset Lower Half}
We set $m=1$, $k=2$ in \Eqref{eq:harmonic_oscillator}.
The initial conditions $q_0$ and $p_0$ are each sampled from a Gaussian distribution with mean $\mu_q= 3$, $\mu_p = 0$ and variance $\sigma_q^2= \sigma_p^2 = 0.2$. 
Training data is generated on the interval $t \in [0, \pi/\sqrt{k}]$. 
The resulting dataset is shown in Figure \ref{fig:datasets} (a).

\paragraph{Dataset Left Half}
We set $m=1$, $k=2$ in \Eqref{eq:harmonic_oscillator}.
The initial conditions $q_0$ and $p_0$ are each sampled from a Gaussian distribution with mean $\mu_q= 0$, $\mu_p = 3\sqrt{k}$ and variance $\sigma_q^2= \sigma_p^2 = 0.2$. 
Training data is generated on the interval $t \in [0, \pi/\sqrt{k}]$. 
The resulting dataset is shown in Figure \ref{fig:datasets} (b).

\paragraph{Dataset Three Quarter}
We set $m=1$, $k=2$ in \Eqref{eq:harmonic_oscillator}.
The initial conditions $q_0$ and $p_0$ are each sampled from a Gaussian distribution with mean $\mu_q= 3$, $\mu_p = 0$ and variance $\sigma_q^2= \sigma_p^2 = 0.2$. 
Training data is generated on the interval $t \in [0, 3\pi/2\sqrt{k}]$.
The resulting dataset is shown in Figure \ref{fig:datasets} (c).

\subsection{Pendulum}
\label{ssec:pendulum}
The damped pendulum is described by
\begin{equation}
	\frac{\dd^2 \phi}{\dd t^2} + \omega^2 \sin \phi + \alpha \frac{\dd \phi}{\dd t} = 0
\end{equation}
where $\phi$ is the angle, $\omega$ frequency and $\alpha$ a damping constant.
For our dataset we set $\omega = 2 \pi/12$ and $\alpha=0.1$.
We choose $t\in [0, 10]$ and add zero mean Gaussian noise with variance $\sigma^2 = 0.1$ to each dataset.
The initial conditions are the same as in \citet{yin2021augmenting}, i.e., $\phi_0  \leftarrow r y_0 / || y_0||_2 $ where $y_0 \sim U_{[0,1] \times [0, 1]}$ and $r =  1.3 + \epsilon, \quad \epsilon \sim U_{[0, 1]}$.  $U$ denotes the uniform distribution over the specified set.
The dataset consists of 25 trajectories.
\subsection{Wave}
The damped wave equation is given by
\begin{equation}
	\frac{\partial^2u}{\partial t ^2} - c^2 \Delta u + k \frac{\partial u}{\partial t} = 0
\end{equation}
For the dataset we set $c=330$ and $k=50$ and $t\in[0, 0.0024]$.
The initial conditions are the same as in \citet{yin2021augmenting}, i.e., 
$u_0 \leftarrow \exp\left[\frac{- (x-m_0)^2  - (y-m_1)^2}{\sigma}\right]$ where
$\sigma \sim U_{[10, 100]}$, $m_0, m_1 \sim U_d\{20, 40\}$. $U_d$ denotes the discrete uniform distribution over the denoted interval.
$x, y$ are 64 dimensional square matrices with $x_{ij}= i$ and $y_{ij} = j$. 
The dataset consists of 200 trajectories.

\section{Implementation details}
\label{sec:implementation}
\subsection{Python packages}
In our code we make use of the following packages: 
Matplotlib \citep{hunter2017matplotlib},
NumPy \citep{harris2020array},
JAX \citep{jax2018github},
Numpyro \citep{phan2019composable},
Laplace-torch \citep{daxberger2021laplace},
SciPy \citep{scipy2020},
PyTorch \citep{pytorch2019} and 
Torchdiffeq \citep{chen2018neural}.

\subsection{Comparison to HMC}
\label{ssec:comparison_hmc}
\subsubsection{Details about HMC settings}
For HMC we use an implementation in JAX \citep{jax2018github} as we found the sampling to be very fast for our particular model and easy to integrate in our existing setup.
For our implementation we use the NUTS sampler supplied by JAX.
We use 2000 samples for warm-up and 2000 samples.
We used a normal distribution with zero mean and variance one as a prior for the weights.
Our implementation is based on \citep{dandekar2020bayesian}. 
For this experiment we try to keep the network architecture as simple and small as possible, to keep sampling times reasonably fast.
\subsubsection{Network Architecture}
We used the same architecture for the HMC model and the Laplace model:
\begin{itemize}
	\item   $\text{Linear}(16, 2) \circ \tanh \circ \text{Linear}(2, 16)$.
\end{itemize}
As an ODE solver we use for both models Dopri5(4) with $rtol=atol=1.4e^{-8}$.
For the MAP trained model we use 10000 training iterations.
\textbf{Computational requirements:} We trained the model on CPU requiring a computation time up to 1.5 h.

\subsection{Hamiltonian Neural ODEs}
\label{ssec:sup_hamiltonian_nodes}
The fact that a system is energy conserving cannot only be encoded into the architecture of the vector field but also in the algorithm of the numerical ODE solver.
Here we describe the symplectic Euler method which is a method of order one, but, similar to classic Runge-Kutta methods, higher order methods exist.
A symplectic Euler step of steps size $h$ is given by
\begin{align}
	p_{n+1} &= p_n - h \partial_q H(p_{n+1}, q_n)\\
	q_{n+1} &= q_n + h \partial_p H(p_{n+1}, q_n).
\end{align}
For separable Hamiltonians, like those used in our experiments, the symplectic Euler is an explicit method (i.e. $\partial_q H(p_{n+1}, q_n) = \partial_q V(q_n)$).
We solve all required ODEs using Euler's method for the naive model, and symplectic Euler for the Hamiltonian neural ODEs.
We set the batch size equal to the dataset size of 16, and use 5000 iterations for training (we trained on CPU).
Each dataset consists of 16 trajectories with slightly different initial conditions.
For our implementation we use code provided by \citep{zhong2020symplectic} and base our architecture on the architecture proposed in this work.
We reduced the network size slightly as this facilitated faster training and using the Laplace approximation on the entire network.
\subsubsection{Neural Network Architectures}
\label{ssec:architecture}
For the experiments on the harmonic oscillator datasets the following architectures were used:
\begin{itemize}
	\item \emph{naive}:   $\text{Linear}(256, 2) \circ \tanh \circ \text{Linear}(2, 256)$
	\item \emph{separable}: $V = T = \text{Linear}(128, 1, bias=false) \circ \tanh \circ \text{Linear}(1, 128)$
	\item \emph{constrained}: $V=\text{Linear}(128, 1, bias=false) \circ \tanh \circ \text{Linear}(1, 128)$
\end{itemize}
\subsubsection{Training of Hamiltonian Neural ODEs}
\label{ssec:sup_training_hnods}
Training of Hamiltonian neural ODEs can be unstable, especially if sufficient structure is not provided (similar observations have been made by \citet{zhong2020symplectic}).
Potentially part of the issue is that Hamiltonian neural networks contain derivatives of neural networks.
These derivatives also change the architecture of the neural network---e.g., a network with two linear layers and a \textit{tanh}-activation has a $1/\cosh^2$-activation after taking the derivative.
Furthermore, different activation functions lead to different extrapolation properties of neural networks \citep{xu2021how}.
That is why the extrapolations of the vector field far away from the data look different for the naive model and the Hamiltonian models.
\textbf{Computational requirements:} We trained the model on CPU requiring a computation time up to 8 h per run.

\subsection{Augmented parametric models}
\subsubsection{Pendulum}
We use the same neural network architecture for the parametric model and the plain neural ODE model:
\begin{itemize}
	\item   $\text{Linear}(32, 2) \circ \tanh \circ \text{Linear}(32, 32) \circ \tanh \circ \text{Linear}(2, 32)$.
\end{itemize}
We use 10000 epochs for training.  
For our implementation we use the code base provided by \citet{yin2021augmenting} and the architecture suggested in this work. 
\textbf{Computational requirements:} We trained the model on CPU requiring a computation time up to 5 h per run.
\subsubsection{Wave}
We use the following neural network architecture for the parametric model:
\begin{itemize}
	\item   $\text{Conv2d}(16, 2, \text{kernel\_size}=3, \text{padding}=1) \circ \text{ReLU}
	\circ \text{Conv2d}(16, 16, \text{kernel\_size}=3, \text{padding}=1, \text{bias}=false)	\circ \text{ReLU} 
	\circ \text{Conv2d}(16, 2, \text{kernel\_size}=3, \text{padding}=1, \text{bias}=false)$.
\end{itemize}
We use 1000 epochs for training. 
For our implementation we use the code base provided by \citet{yin2021augmenting}.
We use the same architecture as in the paper, but removed the batch norm as this provided better results for us.
\textbf{Computational requirements:} We trained the model on GPU (NVIDIA A100 Tensor Core GPU) requiring a computation time up to 12 h per run.

\section{Additional Experimental Results}
\label{sec:sup_experiments}
\subsection{Sampling from the Laplace posterior}
\label{sec:laplace_sample}
\begin{figure}
	\includegraphics[]{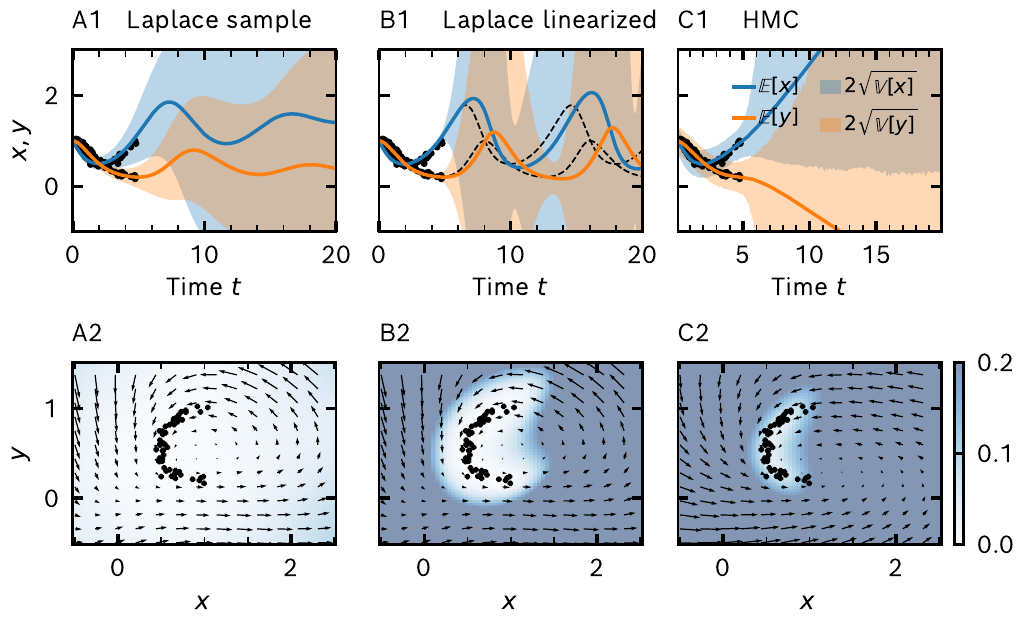}
	\caption{
		(A1--A2) Laplace approach using sampling, (B1--B2) Laplace approach using linearization, (C1--C2) HMC.
	}
	\label{sfig:sample}
\end{figure}

As described in the main text, one option is to sample from the weight posterior. 
We compare a sampling based approach, the linearization approach and HMC on the Lotka Volterra dataset (see Figure \ref{sfig:sample}).

\subsection{HMC on pendulum dataset}
\label{sec:sup_exp_hmc}
\begin{figure}
	\includegraphics[]{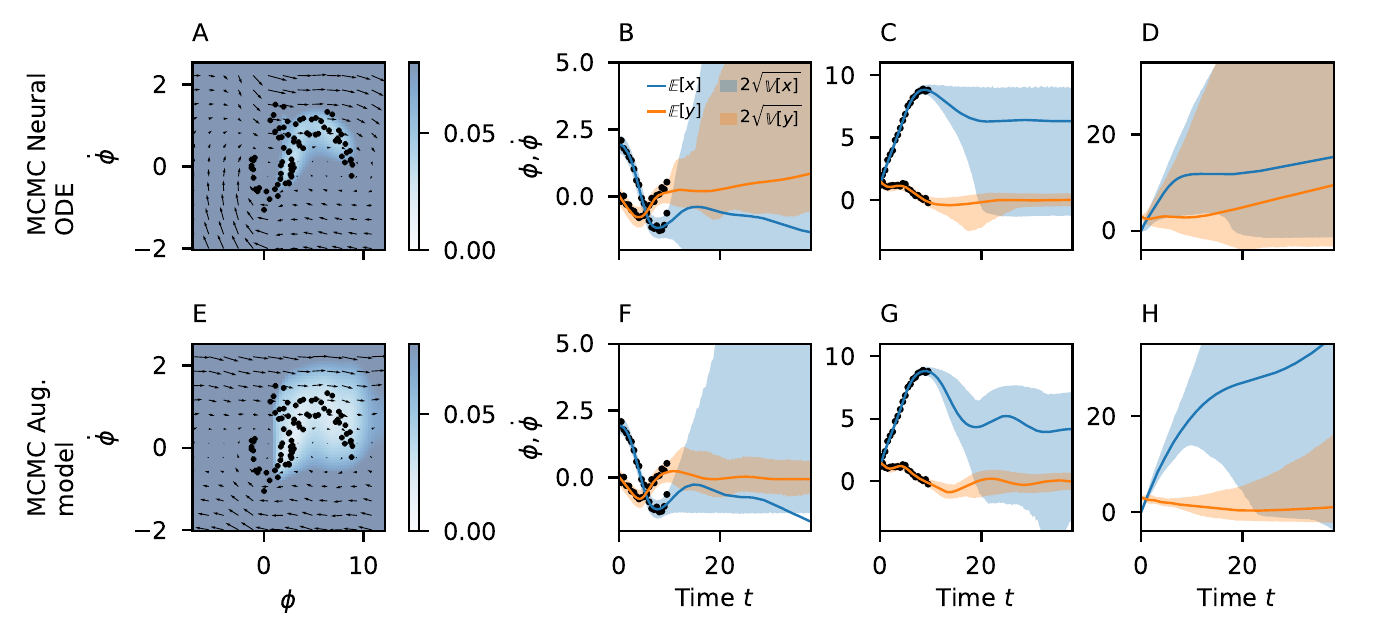}
	\caption{HMC on pendulum dataset. (A--D) Using a plain neural ODE model. (E-H) Using an augmented neural ODE model}
	\label{sfig:mcmc_pendulum}
\end{figure}

We also run a Bayesian neural network using HMC on the damped pendulum dataset from Section \ref{ssec:pendulum}.
However, we use a significantly smaller architecture as for the experiments in the main text, as the HMC sampling becomes too slow if the model has too many weights.
Therefore, we use the same architecture as for the Lotka Volterra experiment (see Section \ref{ssec:comparison_hmc}).
Since the smaller network size is not as expressive as the large network, we reduced the dataset size from 25 samples to 4.
This smaller dataset also leads to a speedup of the HMC approach.
The results for both the plain model and the augmented model are shown in Figure \ref{sfig:mcmc_pendulum}.
We find that as for the experiments in the main text, the vector field of the augmented model has a larger area of low uncertainty than the plain neural ODE model.
Similarly, we observe that the extrapolation behavior augmented model is better (see Figure \ref{sfig:mcmc_pendulum}).
Increasing the dataset would again improve extrapolation behavior even further, but also requires a network large enough to be expressive enough to capture all the data.

\clearpage
\subsection{Hamiltonian Neural ODEs}
\label{sec:sup_exp_hamiltonian}
\begin{figure}[H]
	\centering
	\subfloat[]{
		\includegraphics[width=0.4\textwidth]
		{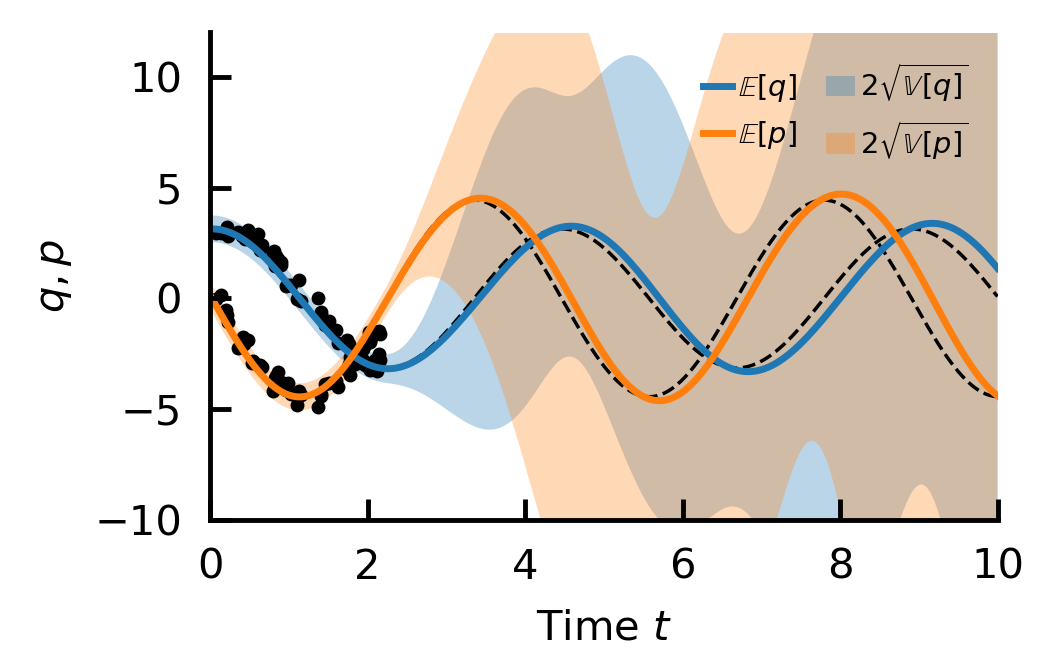}
	}
	\subfloat[]{
		\includegraphics[width=0.4\textwidth]
		{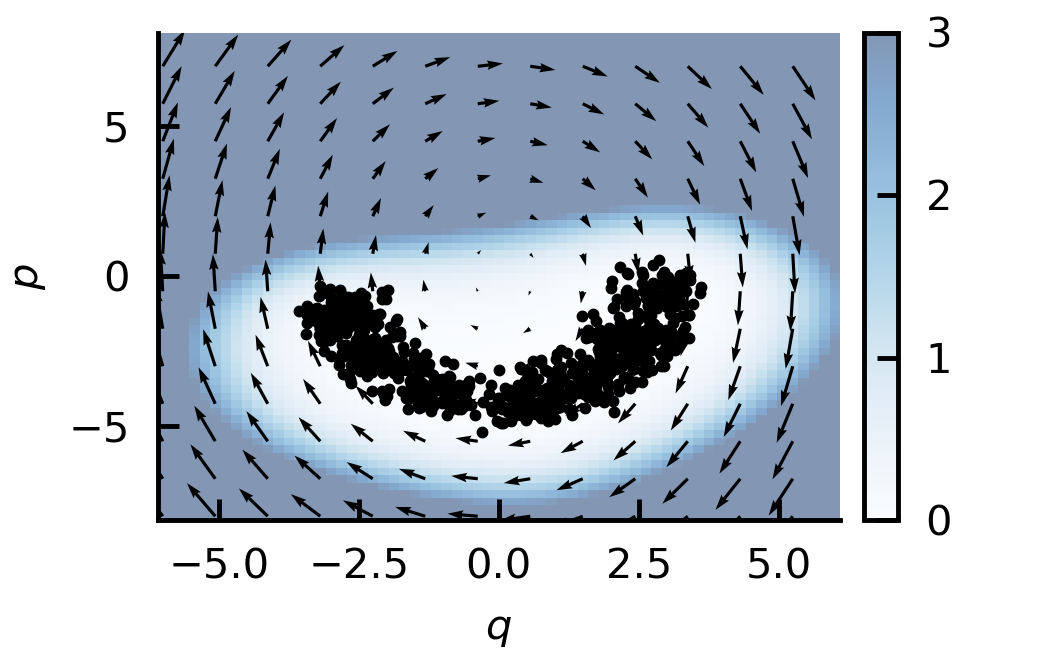}
	}

	\subfloat[]{
		\includegraphics[width=0.4\textwidth]
		{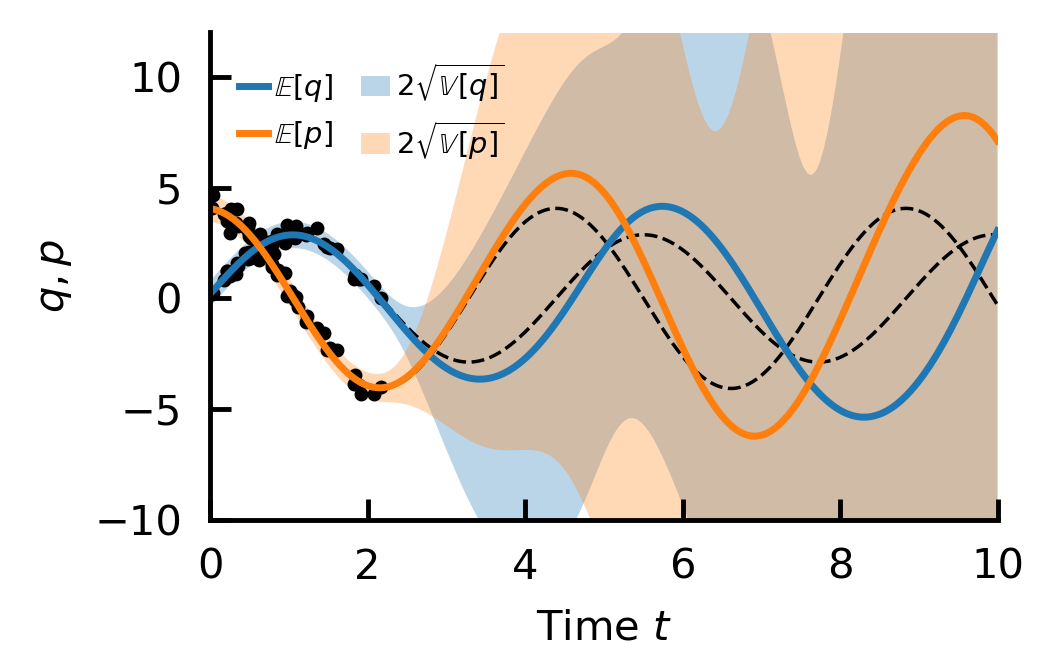}
	}
	\subfloat[]{
		\includegraphics[width=0.4\textwidth]
		{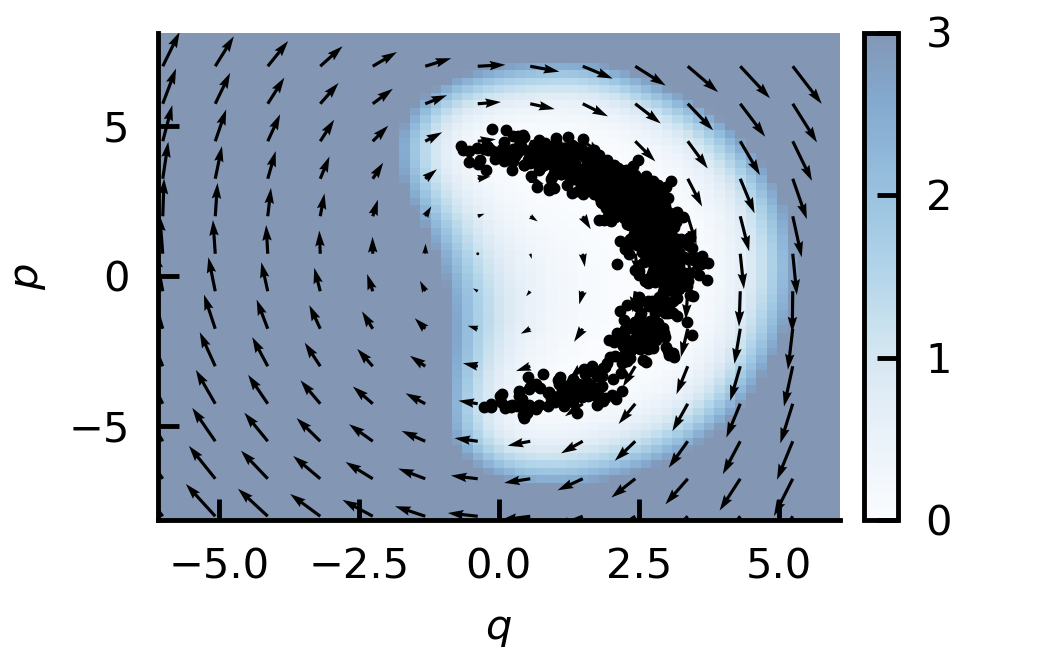}
	}
	
	\subfloat[]{
		\includegraphics[width=0.4\textwidth]
		{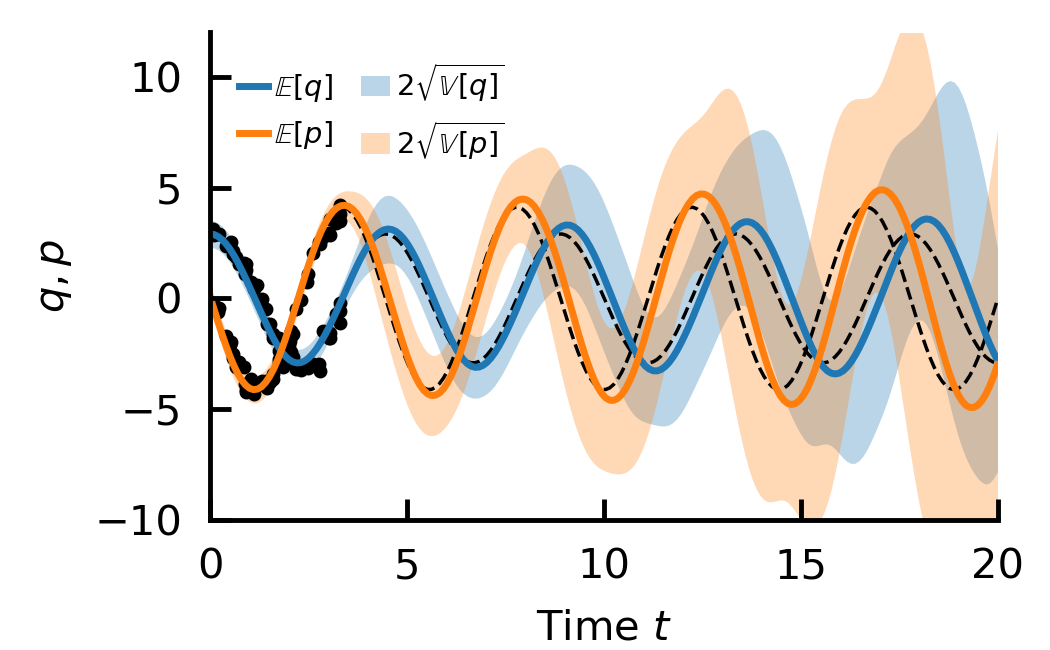}
	}
	\subfloat[]{
		\includegraphics[width=0.4\textwidth]
		{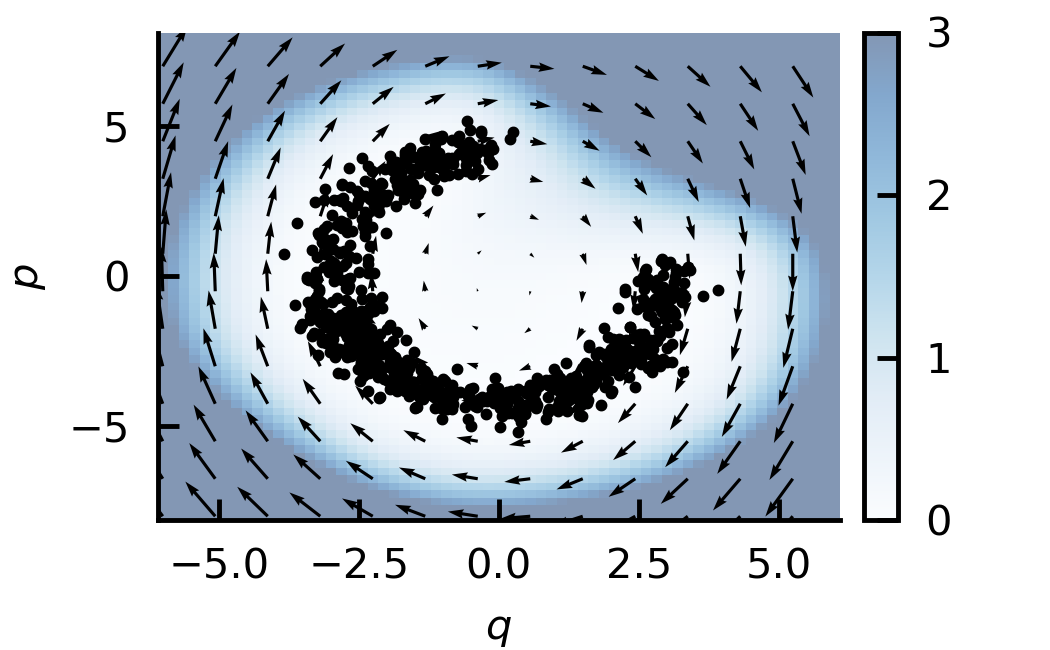}
	}

	\caption{Naive network trained on harmonic oscillator using Euler's method. Training on \emph{dataset-lower-half} (a--b), \emph{full-cycle-dataset} (c--d), \emph{dataset-three-quarter} (e--f).}
	\label{fig:naive_euler}
\end{figure}

\begin{figure}
	\centering
	\subfloat[]{
		\includegraphics[width=0.4\textwidth]
		{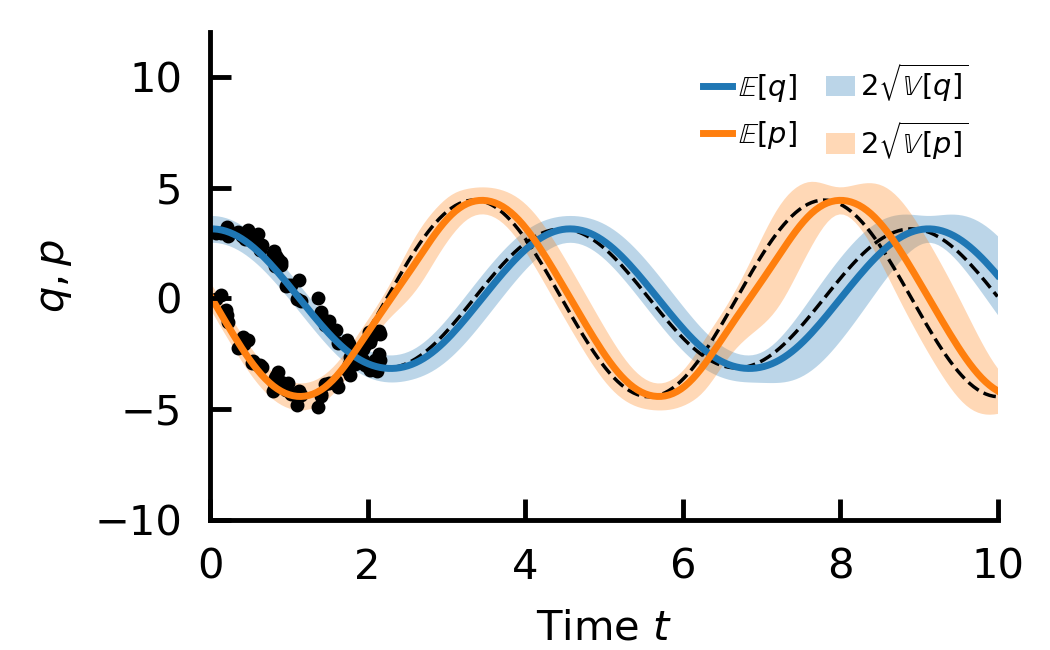}
	}
	\subfloat[]{
		\includegraphics[width=0.4\textwidth]
		{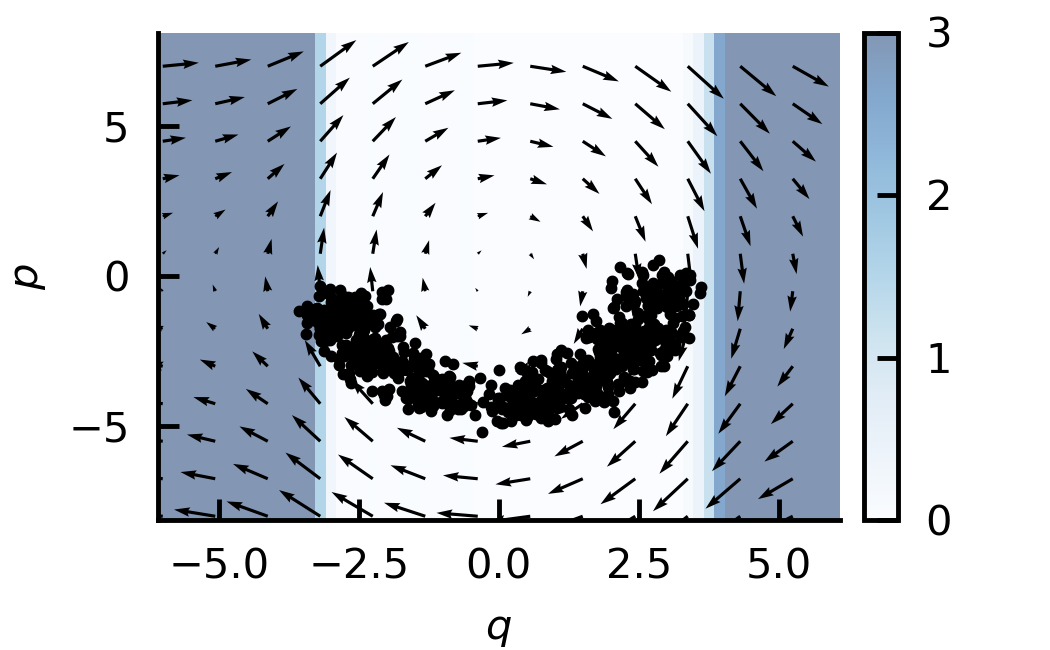}
	}
	
	\subfloat[]{
		\includegraphics[width=0.4\textwidth]
		{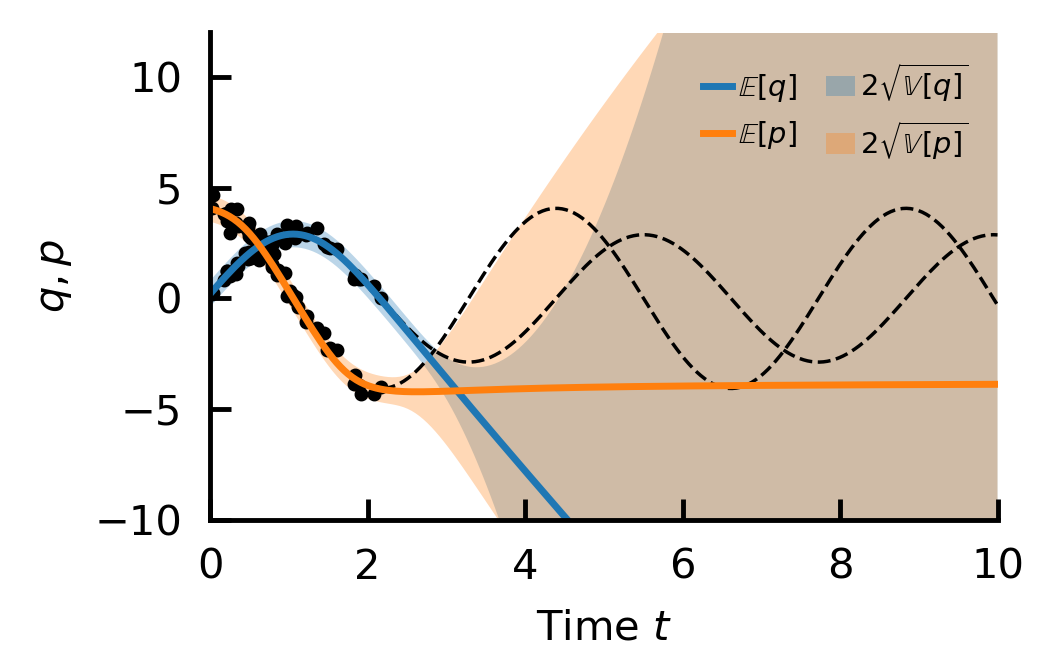}
	}
	\subfloat[]{
		\includegraphics[width=0.4\textwidth]
		{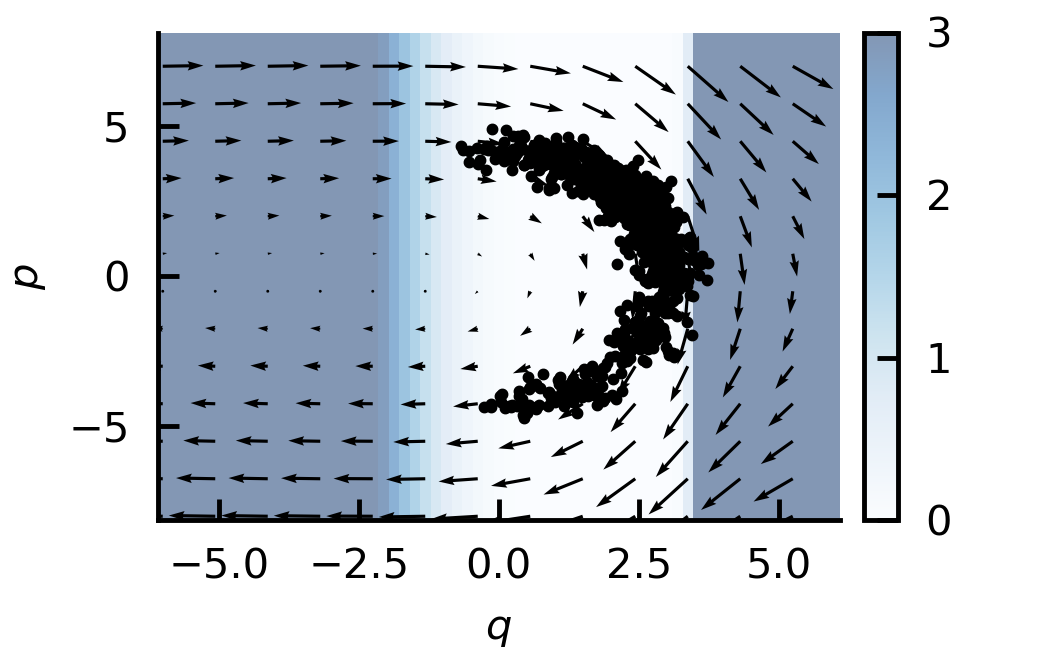}
	}
	
	\subfloat[]{
		\includegraphics[width=0.4\textwidth]
		{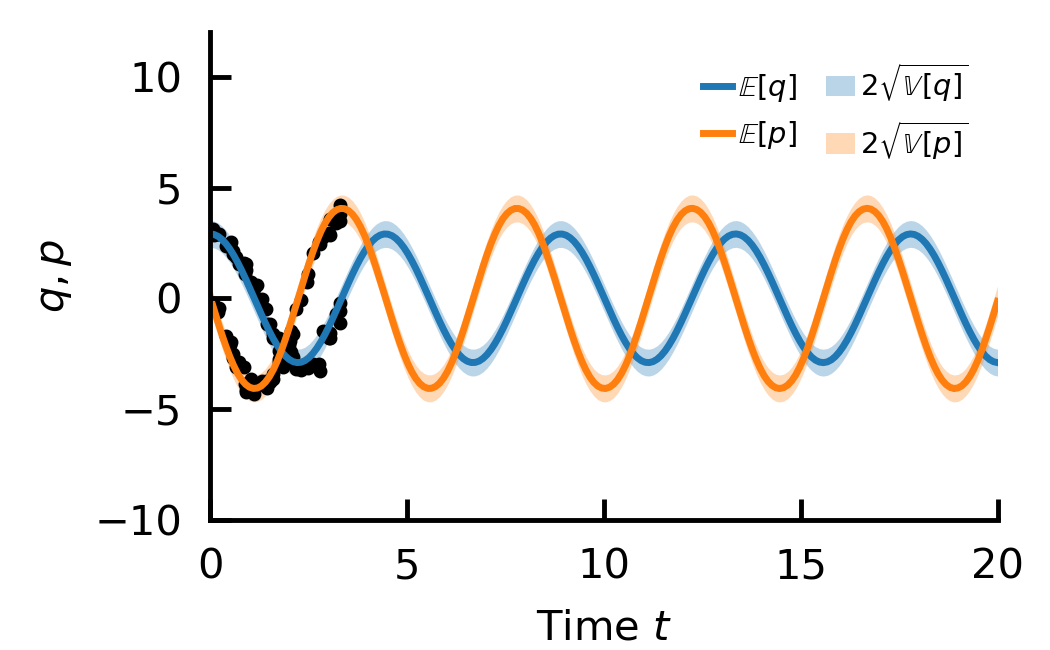}
	}
	\subfloat[]{
		\includegraphics[width=0.4\textwidth]
		{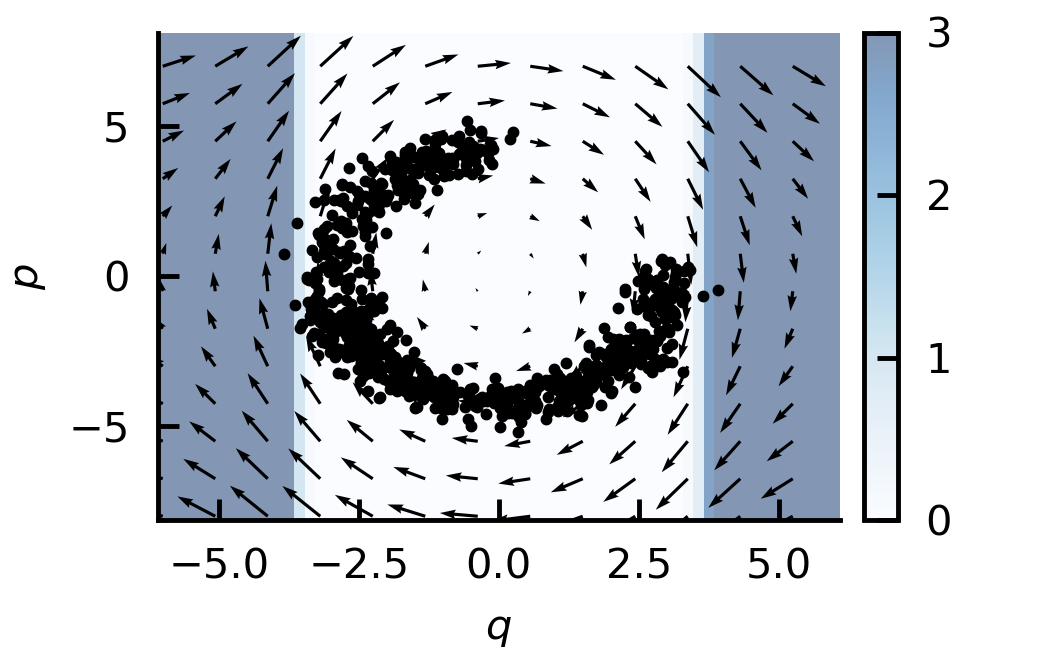}
	}

	\caption{Constrained network trained on harmonic oscillator using symplectic Euler. Training on \emph{dataset-lower-half} (a--b), \emph{dataset-left-half} (c--d), \emph{dataset-three-quarter} (e--f).}
	\label{fig:constrained_symplectic}
\end{figure}

\begin{figure}
	\centering
	\subfloat[]{
		\includegraphics[width=0.4\textwidth]
		{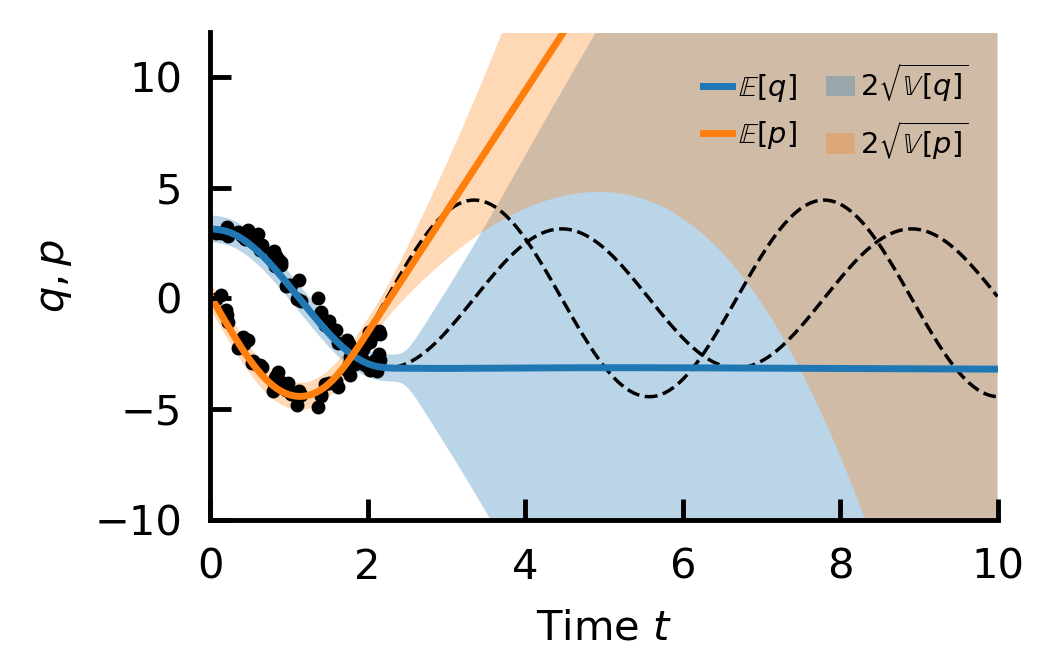}
	}
	\subfloat[]{
		\includegraphics[width=0.4\textwidth]
		{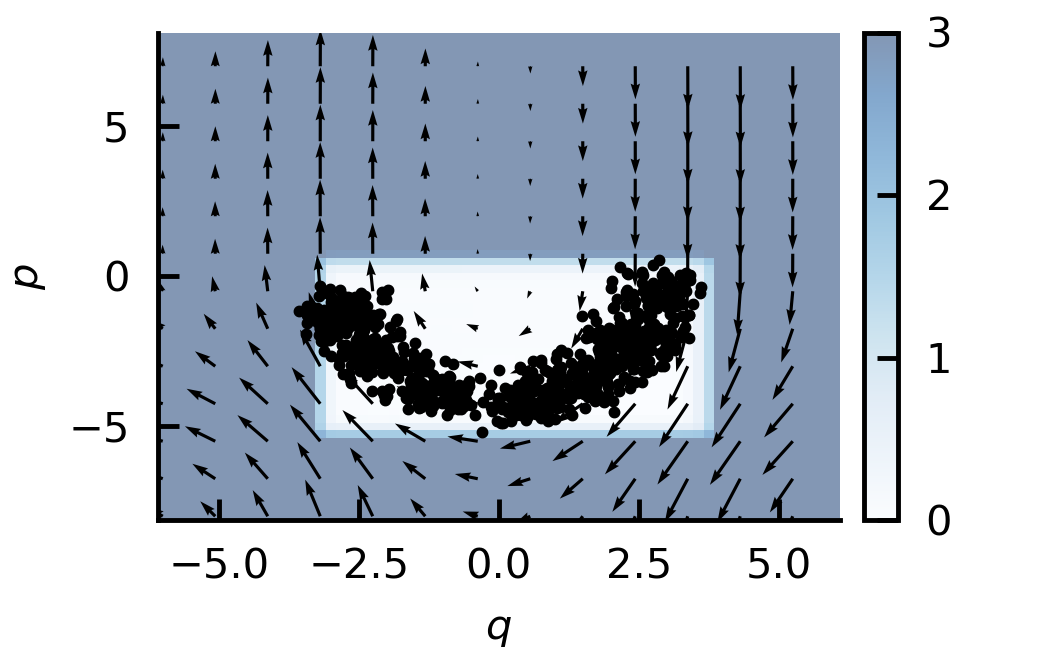}
	}
	
	\subfloat[]{
		\includegraphics[width=0.4\textwidth]
		{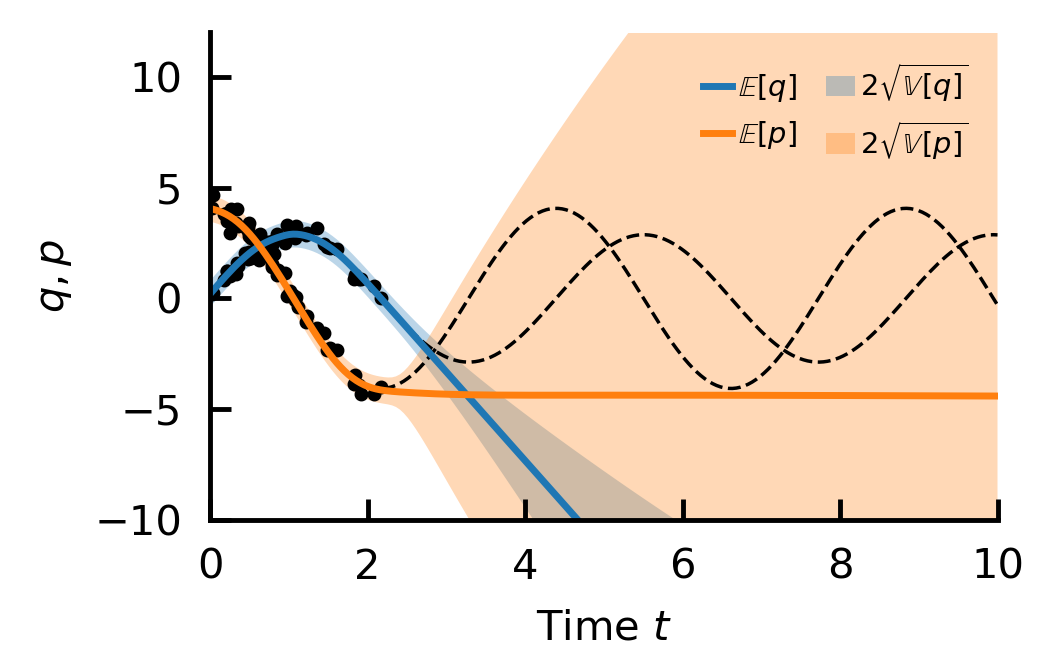}
	}
	\subfloat[]{
		\includegraphics[width=0.4\textwidth]
		{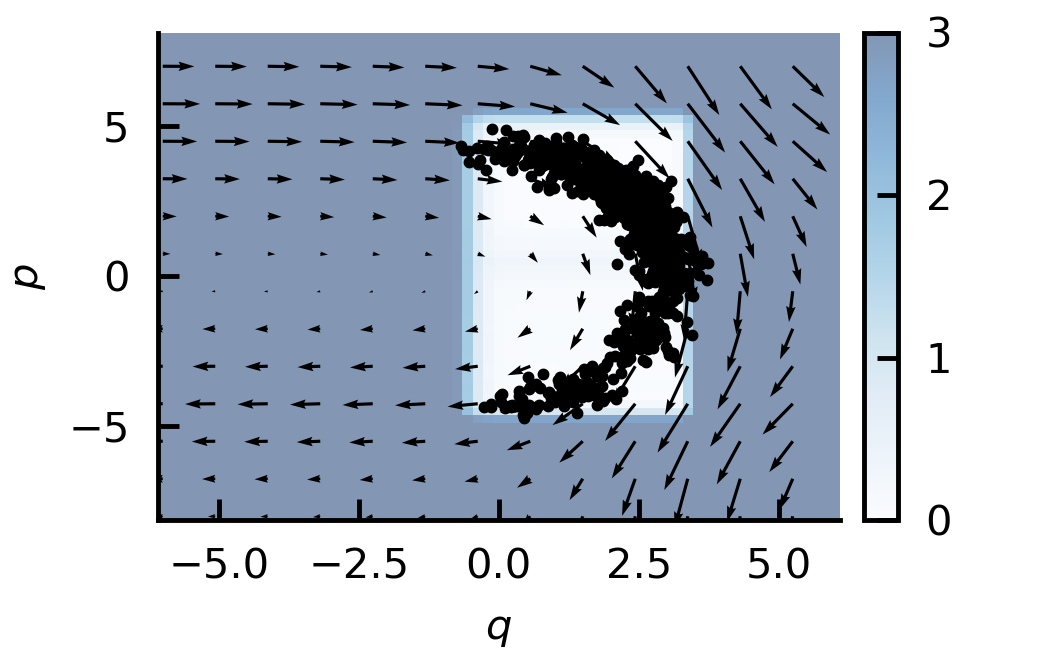}
	}
	
	\subfloat[]{
		\includegraphics[width=0.4\textwidth]
		{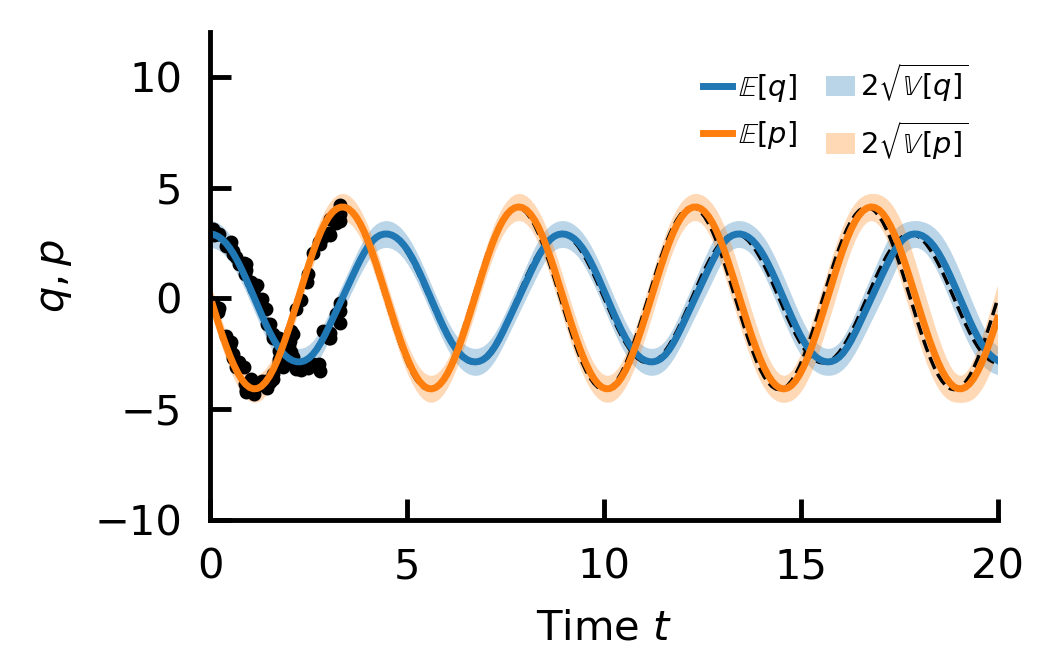}
	}
	\subfloat[]{
		\includegraphics[width=0.4\textwidth]
		{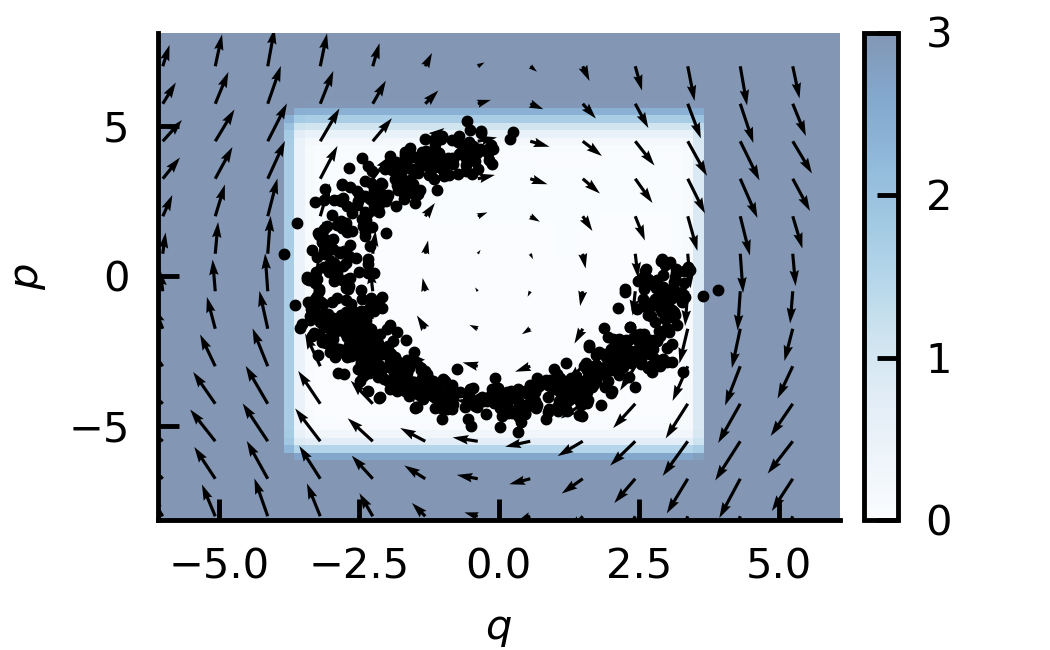}
	}
	
	\caption{Separable network trained on harmonic oscillator using symplectic Euler. Training on \emph{dataset-lower-half} (a--b), \emph{dataset-left-half} (c--d), \emph{dataset-three-quarter} (e--f).}
	\label{fig:separable_symplectic}
\end{figure}
\clearpage
\subsection{Additional wave results}
\label{sec:sup_wave_exp}
\begin{figure}[H]
	\includegraphics[width=\textwidth]{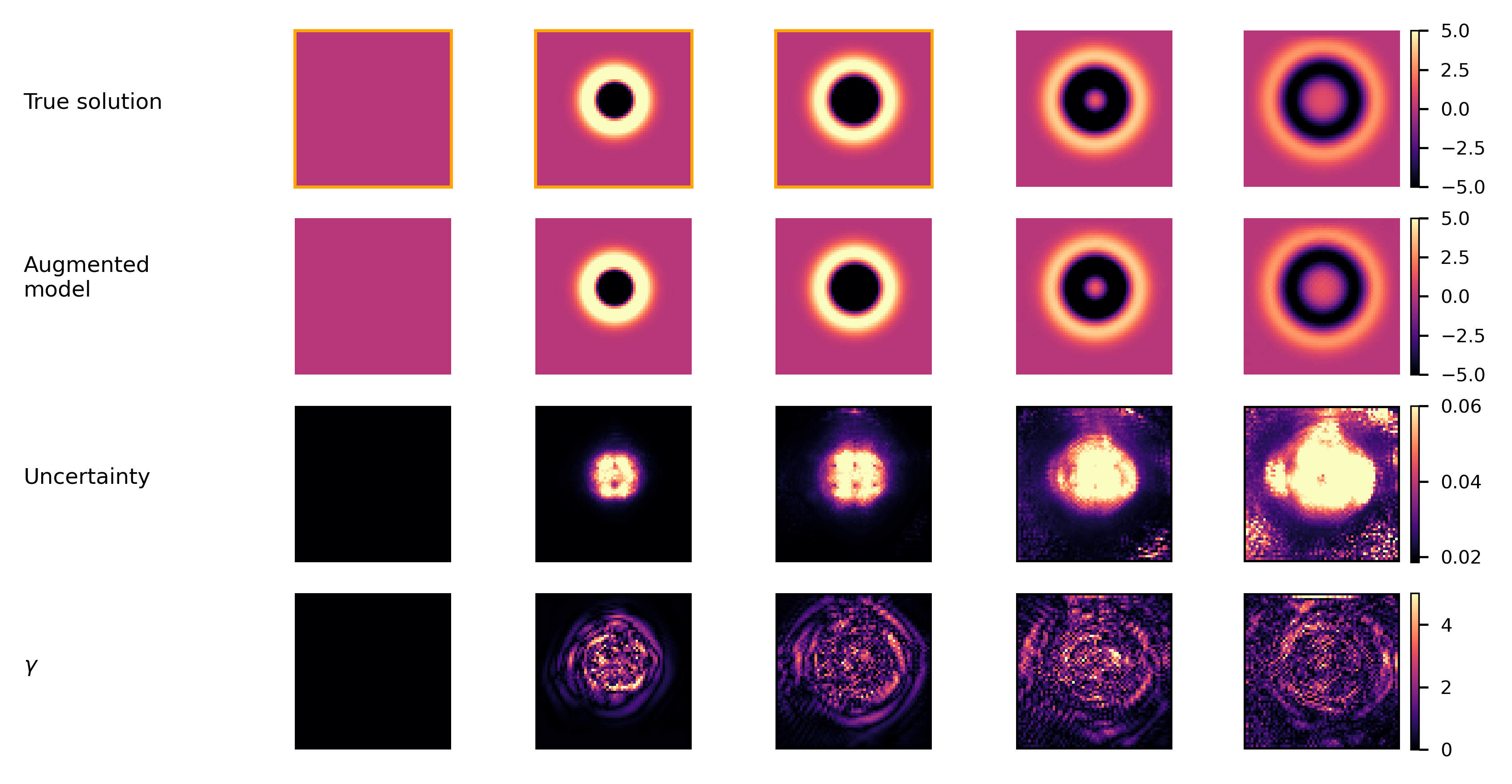}
	\caption{
		\textit{Laplace approximation applied to wave PDE.}
		Training an augmented parametric model on the damped wave equation dataset (Figure shows result for $\dd u/ \dd t$).
		First three images (marked by orange frame) are part of the training data.}
\end{figure}
\begin{figure}[H]
	\includegraphics[width=\textwidth]{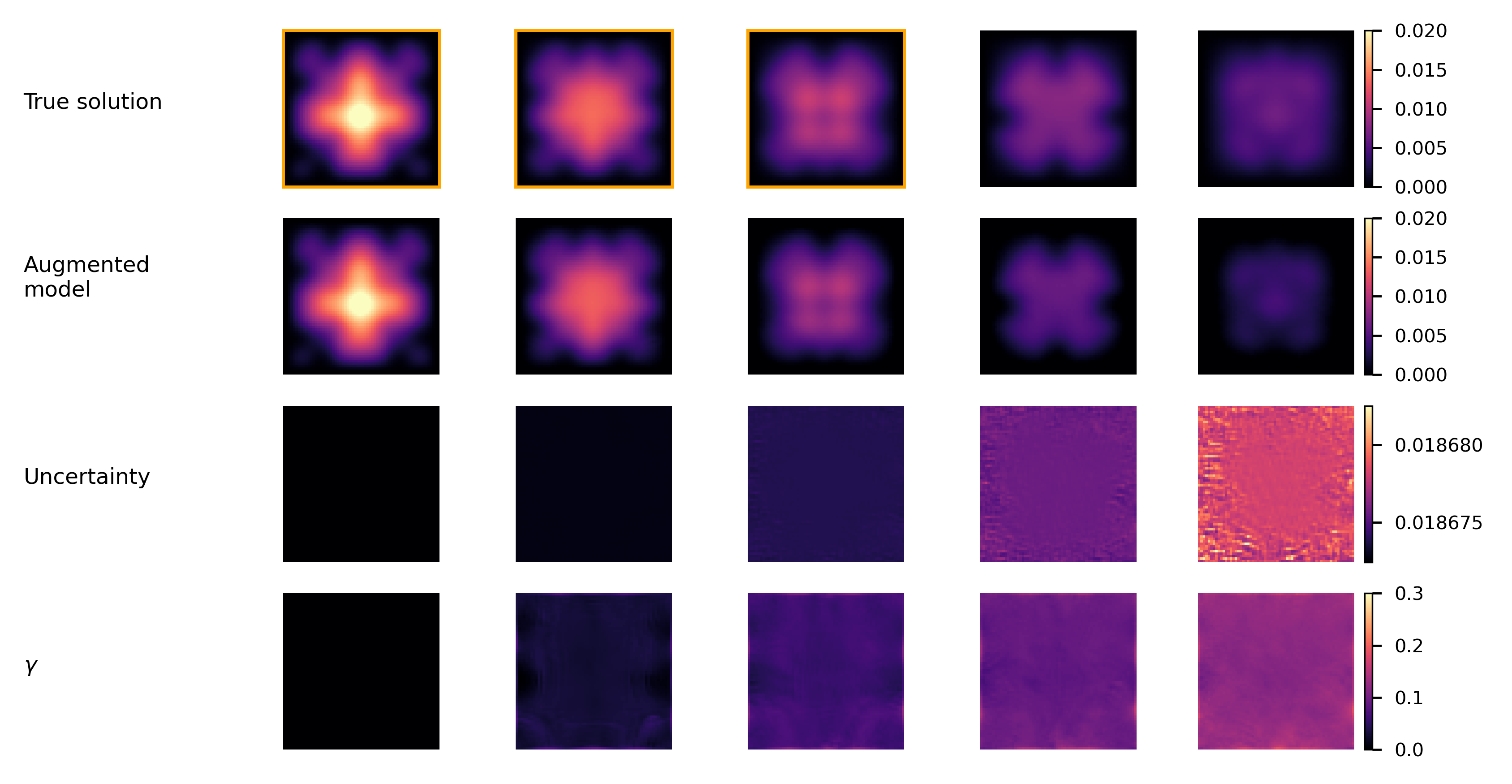}
	\caption{
		\textit{Laplace approximation applied to wave PDE.}
		Training an augmented parametric model on the damped wave equation dataset with a different initial condition (Figure shows result for $u$).
		First three images (marked by orange frame) are part of the training data.
	}
\end{figure}
\begin{figure}[H]
	\includegraphics[width=\textwidth]{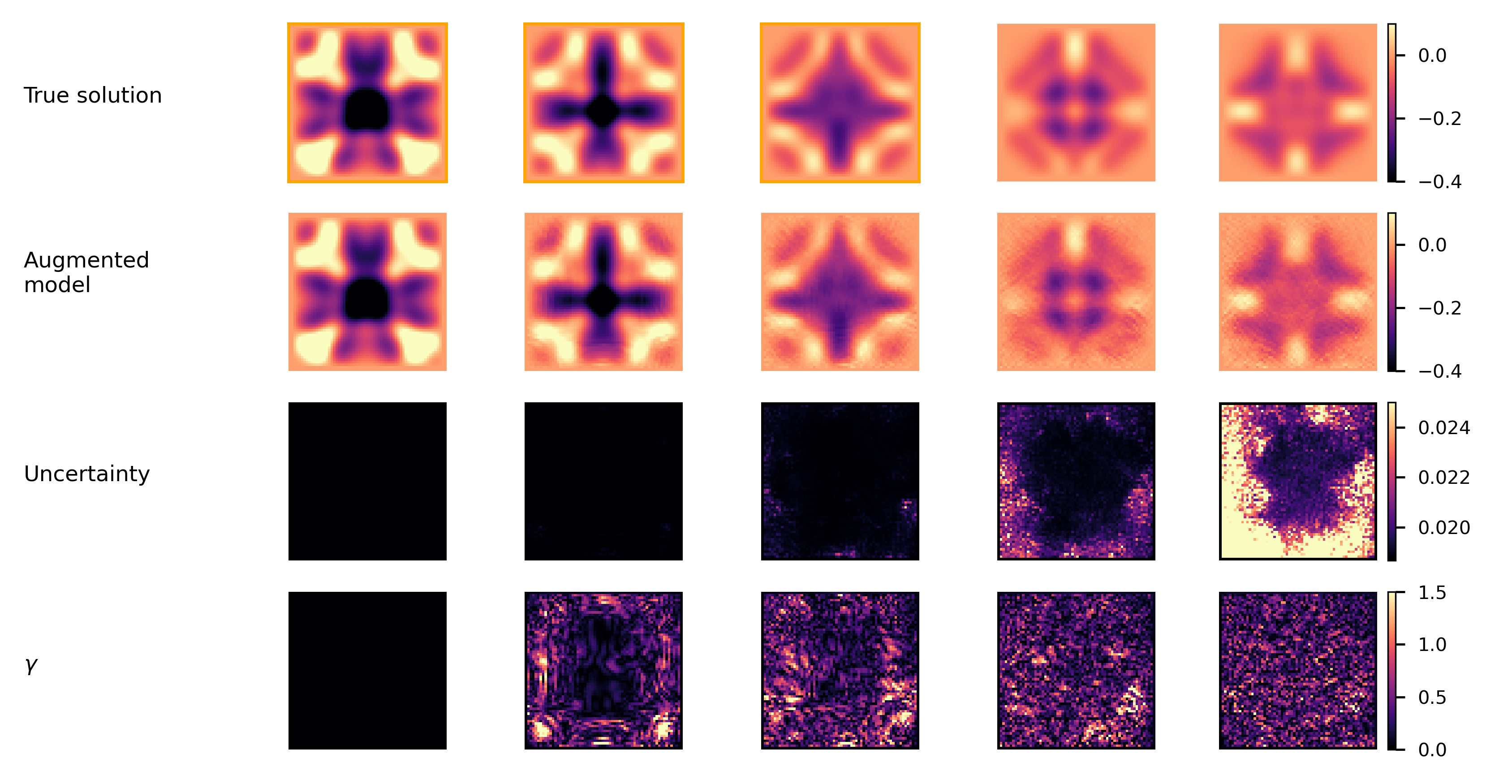}
	\caption{
		\textit{Laplace approximation applied to wave PDE.}
		Training an augmented parametric model on the damped wave equation dataset with a different initial condition (Figure shows result for $\dd u/ \dd t$).
		First three images (marked by orange frame) are part of the training data.
	}
\end{figure}

\end{document}